%% file: paper_full.tex
\documentclass[11pt]{article}

\makeatletter
\IfFileExists{sourceserifpro.sty}{%
  \usepackage{paper}%
}{%
  \usepackage[T1]{fontenc}
  \usepackage[utf8]{inputenc}
  \usepackage{lmodern}
  \usepackage{amsmath,amssymb}
  \usepackage[top=1in, left=1.3in, right=1.3in, bottom=1.2in, footskip=0.7in]{geometry}
  \usepackage[onehalfspacing]{setspace}
  \usepackage{microtype}
  \usepackage{xcolor}
  \usepackage{natbib}
  \setcitestyle{aysep={}}
  
  \setlength{\bibsep}{0pt}
  \setlength{\bibhang}{\parindent}
  \renewcommand\section{\@startsection{section}{1}{\z@}%
    {-3.5ex \@plus -1ex \@minus -.2ex}{2.3ex \@plus.2ex}%
    {\normalfont\large\bfseries}}
  \renewcommand\subsection{\@startsection{subsection}{2}{\z@}%
    {-3.25ex\@plus -1ex \@minus -.2ex}{1.5ex \@plus .2ex}%
    {\normalfont\normalsize\bfseries}}
  \renewcommand\paragraph{\@startsection{paragraph}{4}{\z@}%
    {3.25ex \@plus1ex \@minus.2ex}{-1em}{\normalfont\normalsize\itshape}}
  \renewcommand{\thesection}{\arabic{section}}
  \AtBeginDocument{\frenchspacing}
}
\makeatother
\usepackage{graphicx}
\usepackage{booktabs}
\usepackage{multirow}
\usepackage{url}
\usepackage{hyperref}
\hypersetup{hidelinks,pdftitle={The Value of Personalization in Large Language Model Responses}}

\newcommand{\sigmoid}{\operatorname{logit}^{-1}}

\input{tables/steering_numbers.tex}
\input{tables/prose_numbers.tex}

\title{The Value of Personalization in Large Language Model Responses}
\author{Andrey Fradkin$^{1}$, Yan (Stella) Si$^{2}$, Georgios Zervas$^{1,2}$\\[2pt]
  {\small $^{1}$Questrom School of Business, Boston University}\\[-2pt]
  {\small $^{2}$Faculty of Computing \& Data Sciences, Boston University}}
\date{\today{}}

\begin{document}

\maketitle

\begin{abstract}
\noindent
We measure the benefits of matching large language model (LLM) responses to users and prompts. To do so, we estimate user preferences and use them to evaluate routing strategies. Against always serving the single model with the highest average utility, a router using both the prompt and the user's preferences attains a \UserPromptWinRounded{} predicted win rate, versus \PromptWinRounded{} for a router using the prompt alone, so most of the gain comes from knowing the user. These routers choose among already-generated responses; picking a model before generation performs worse, since it requires predicting each model's response. Steering the baseline model by stating a user's preferences in its prompt performs about as well as personalized routing, and combining routing with steering performs even better. The router that knows the user achieves these gains while selecting models with lower average benchmark scores. Because users have heterogeneous preferences over responses, routing across diverse models creates value that benchmark rankings miss.
\end{abstract}

\pagebreak

\section{Introduction}\label{s:introduction}

A common view of progress in AI is that scaling training data, compute, and inference improves performance on benchmarks without limit. This view assumes that one dimension of intelligence explains most of a model's value. Yet LLM responses also differ in style, including length, tone, and formatting, which are distinct from the intelligence of a model. If users value these characteristics differently across tasks, no single model is best for everyone. In this paper, we measure the benefits from routing requests across models based on users' heterogeneous preferences for response characteristics.

We begin by estimating a random-coefficients discrete-choice model
in which utility depends on the judged accuracy of a response, model identity and capabilities, response style, the prompt, and a user's latent tastes. The estimation uses preference data from Arena, a platform that presents users with anonymized
responses from two randomly assigned models and asks which they prefer
\citep{chiang2024arena}. Accuracy is the most valued attribute of a response. Yet how much accuracy matters depends on the prompt, and users differ widely in how much they value it. Style also matters, and users differ in their tastes for features such as tables and length. Our discrete-choice model allows us to simulate routing strategies without running live experiments.

We propose a simple framework for evaluating the gains from routing. As an illustration, consider a user who usually likes detailed explanations but asks for a short explanation today. Past choices can reveal the usual preference, while the prompt can reveal what the user wants this time. The platform could route the request to a model that tends to answer briefly, it could steer the usual model by asking it to give the explanation in one sentence, or it could do both. The platform can also sample responses from several models and serve the best one. The gains from using these routing strategies are a function of user preferences, the platform's knowledge about those preferences, and the platform's knowledge about responses across candidate models and instructions.

We first compare three routing strategies against always serving the best single model in our sample (Gemini 2.5 Pro). A prompt router serves the baseline model on \PromptKeepsBaseline{} of prompts and switches models on the remaining \PromptSwitches{}, and wins \PromptWin{} of the time against the baseline under the model. A router that knows only the user assigns each user one model for all prompts: \PersonalBestDiffers{} of users receive a different model from the one that is best for the average user; this router attains a \KnownUserWin{} win rate on held-out prompts.

We then consider a user-by-prompt router that uses both the prompt and each user's taste vector. In this case, no model dominates the ten-model candidate set: the leading model accounts for only \LargestShare{} of user-prompt choices, and the top two models together receive \TopTwoShare{}, leaving more than half to eight other models. Knowing each user's tastes raises the win rate to \UserPromptWin{}. Knowing the user, beyond the prompt, accounts for \PersonalGainShare{} of the total utility gain. The user-by-prompt router achieves this gain while selecting models with lower benchmark scores. Its selections average \OracleBenchmark{} on the Artificial Analysis index, compared with \FixedBaselineBenchmark{} for the fixed Gemini 2.5 Pro baseline and \PromptBenchmark{} for the prompt router.

Next, we measure how much of this utility gain a platform can achieve by learning tastes from pairwise choices. We simulate \FeedbackRounds{} rounds of pairwise feedback and use a Bayesian approach to learn the preference vector. This approach closes \FortyChoicesClosed{} of the utility gap between the prompt router and the user-by-prompt router. Our results on learning preferences for responses are closely related to recent methods that represent users by latent variables or ideal points in a shared preference space \citep{poddar2024variational,chen2025pal}, as well as \citet{shenfeld2025reward}, who factorize user-specific rewards into shared base functions and use active learning to infer a new user's weights from a small number of pairwise comparisons.

We also test an intervention that requires no model fine-tuning but does require knowing the user's preferences: a
natural-language preamble that instructs the model to match the user's tastes.
On a \SteeringPrompts-prompt evaluation subset, steering the baseline model (Gemini 2.5 Pro) wins against its unsteered panel response \SteeredPanelWin{} of the time, compared with \SteeringSubsetOracleWin{} for the user-by-prompt router across all ten models' original responses on the same prompts and users, and steering raises the intended user's utility for every model we steer. This comparison depends on how users value links, which account for \verbLinkSh{}\% of the steering gain over a fresh unsteered response. Routing over the original and steered responses of the six models we steer wins \CombinedSteeringWin{} of the time.

Our main router chooses after generation: it generates one response from each of ten models and then selects among them, which raises inference costs and latency. As an alternative, we evaluate routers that choose a model before generation, which \citet{hu2024routerbench} call predictive routing. We benchmark these against the same baseline, always using the best model in the sample, on held-out prompts and users, where the user-by-prompt router attains a \HeldoutPersonalWin{} win rate and the prompt router \HeldoutPromptWin{}. In one approach, we predict each model's response characteristics, including its judged accuracy, from a numerical representation (embedding) of the prompt and score them through the choice model. This prompt-based router that chooses before generation attains a win rate of only \ExantePromptWin{}, essentially a coin flip against the baseline: response characteristics are hard to predict before generation.

In another approach, we build a router that uses a taste vector learned from \FeedbackRounds{} of the user's past pairwise choices to pick one best model without generating responses from all models; it attains a \ExanteUserWin{} win rate and \ExanteUserGainShare{} of the user-by-prompt router's gain in mean utility. Combining the two approaches raises the win rate to \ExanteCombinedWin{} and the share of that utility gain to \ExanteCombinedGainShare{}. One reason for the shortfall is sampling variance. The utility of the same model's response varies across draws as much as the prompt-specific edge that a router exploits.

One concern with studies like ours is that the data used becomes outdated very quickly. The latest models are much better than those in our sample. Rather than focusing on the exact routing gains, we view our main contribution as the framework for routing and its combination with an estimated demand model. Any platform with data on user feedback about responses can implement our strategy.

The closest paper to ours is \citet{caoui2025consumer}, who uses Arena choices to
estimate demand for differentiated LLMs and decomposes changes in consumer
surplus into improvements in intelligence and other model attributes. We use a similar demand framework but shift the counterfactual from measuring
market-wide quality growth to assigning responses across models and users. Our
contribution is to quantify separately the value of prompt-level routing and
user-level personalization in a choice model that accounts for judged
response accuracy, and to evaluate mechanisms for learning and acting on individual
tastes. 

In the LLM routing literature, \citet{hu2024routerbench} introduce
RouterBench to evaluate prompt-level routers on the trade-off between inference
cost and a single benchmark quality metric across tasks. By contrast, we
abstract from inference costs and evaluate routing with an estimated demand
model of heterogeneous human preferences over accuracy, style, and models,
allowing us to separate user-level personalization from prompt-level routing
and compare both with steering a single model. The paper also complements work documenting pluralistic preferences in
human feedback \citep{kirk2024prism} and divergence between aggregate and
individual LLM rankings \citep{garbacea2026personalized}. Our findings are also
consistent with \citet{demirer2025emerging,demirer2026market}, who document horizontal
and vertical differentiation in the market for LLM inference and show that no
single model dominates across use cases. More broadly, our results provide
empirical content for models of product design and the pricing of LLM services
\citep{bergemann2025economics}.

The paper proceeds as follows. Section~\ref{sec:data} describes the data, response attributes, and accuracy judge. Section~\ref{sec:framework} presents the choice model and identification. Section~\ref{sec:estimates} reports the estimates. Section~\ref{sec:results} measures the gains from routing, learning tastes from feedback, choosing a model before generation, and steering. Section~\ref{sec:conclusion} concludes and discusses limitations. The appendices give the data construction, the full coefficient tables and identification diagnostics, and the details of every simulation.

\section{Data}
\label{sec:data}

Arena, originally introduced as Chatbot Arena and also known as LMArena, is an online platform for evaluating large language models through human preference comparisons \citep{chiang2024arena}. Users submit prompts and receive side-by-side responses from two randomly assigned models, shown anonymously. Arena calls these comparisons ``battles.'' Users can select either response, declare a tie, or judge both responses to be bad. Voting is voluntary; if users vote, Arena then reveals the model identities. Arena aggregates these votes into a public leaderboard. The data thus reflect users' decisions both to use Arena and to vote.

We pool two public data releases covering June--August 2024 and April--July 2025. Each battle record contains the user's prompt, the two models' responses, their model identities, and the user's vote. We restrict the sample to English-language battles and drop incomplete observations.\footnote{The releases are \texttt{arena-human-preference-100k} and \texttt{arena-human-preference-140k}. We exclude battles without a usable vote, with an empty response, or without complete response features, model attributes, and prompt scores, as described below.} We split battles at random into training (\TrainShare) and test (\TestShare) samples, yielding \TrainBattles{} training and \TestBattles{} test battles. We use the training sample to estimate preferences and the test sample to assess predictive fit. We count a tie as half a win for each response and retain ``both bad'' votes as a separate outcome. Table~\ref{tab:estimation-variables} reports descriptive statistics for the estimation sample.

For each response, we construct thirteen features: log token length; counts of ordered lists, unordered lists, headers, and bold spans; indicators for tables, emoji, and links; Flesch reading ease; positive and negative sentiment; and assertive and tentative hedging. We calculate length, formatting, and readability measures directly from each response. We use gpt-oss-20b, an open-weight language model, to classify sentiment and hedging. We standardize length, Flesch reading ease, and the list, header, and bold-span counts using training-sample means and standard deviations; the binary features stay 0/1.

We score every response in the estimation sample for factual accuracy using GLM 5.2, chosen for its low cost when scoring at scale. The judge sees the prompt and response and returns a score from 0 to 10 with a short rationale. The choice model uses the score itself (Section~\ref{sec:framework}). To assess how sensitive the scores are to our choice of judge, we rescore a validation sample with Claude Opus 4.8. When we group scores into low (0--3), mid (4--6), and high (7--10), the two judges agree on an estimated \JudgeBinAgreement{} of responses in the corpus (Appendix~\ref{app:accuracy}).

To capture how preferences vary with the task, we summarize each prompt with three scores. We embed each prompt and use partial least squares (PLS) to find the directions of the embedding that best predict how response features differ between winners and losers. We fit this projection on training battles and apply it to all prompts in our sample. Appendix~\ref{app:data} details these steps.

We measure model intelligence with version 2 of the Artificial Analysis Intelligence Index (July 2025 snapshot): a weighted average of seven benchmark scores (MMLU-Pro, Humanity's Last Exam, GPQA Diamond, MATH-500, AIME 2024, SciCode, and LiveCodeBench), averaged over the available benchmarks when a model lacks some.\footnote{The current version of the index (v5.0.1, November 2026) no longer scores our models, which average \AAIndexMean{} on version 2. On the Epoch Capabilities Index (ECI) as of November 2026, the \ECIModels{} sample models it covers average \ECIMean{}.} We take the scores from Artificial Analysis where it reports them and from model documentation otherwise. The index covers all \EstimatedModels{} models in our sample, and we scale it in units of ten points.\footnote{The Epoch Capabilities Index predicts judged accuracy somewhat better at the model level ($R^2$ of \ECIModelFit{} against \IndexModelFit{}, on the models both indices cover) but is available for only \ECIModels{} of our models.} We also record each model's context window in millions of tokens and an indicator for whether it uses extended chain-of-thought reasoning before responding.

More capable models give more accurate answers on average, but most of the variation in accuracy lies within a model. Among the \TopModels{} models with the most responses, the mean judged score ranges from \TopModelsMinScore{} to \TopModelsMaxScore{} (Appendix Figure~\ref{fig:accbymodel}). Across the \IndexedModels{} models with a benchmark score, the \EstimatedModels{} in the estimation sample and the one we drop for appearing in fewer than \MinModelBattles{} battles, the mean score correlates at \ScoreBenchmarkCorr{} with the benchmark index. Differences between model means account for only \BetweenModelVariance{} of the variance in judged scores. The rest lies among responses from the same model.

\begin{table}[!htbp]
\caption{Descriptive statistics for the Arena estimation sample.}
\label{tab:estimation-variables}
\label{tab:features}
\centering
\small
\input{tables/tab_estimation_variables.tex}
\par\vspace{4pt}
\begin{minipage}[t]{\linewidth}
\footnotesize\input{tables/notes_estimation_variables.tex}
\end{minipage}
\end{table}

To evaluate routing across models, we assemble an additional response panel using prompts sampled from Arena. We select ten models for comparison. For each sampled prompt, we retain any responses from these models already recorded in Arena and request a response from each model whose answer is missing. The complete panel covers \PanelPrompts{} prompts. We measure its response attributes and accuracy using the same procedures as for the estimation data and compute prompt scores using the PLS projection fitted on training battles. Almost all panel prompts come from battles in the estimation sample, most of them training battles, so the routing results evaluate policies on prompts the preference model has seen. Section~\ref{sec:results} lists the candidate models. We also generate responses under personalized instructions to study steering (Section~\ref{sec:steering}) and draw repeated responses to the same prompts to study sampling variation (Section~\ref{sec:noise}).

\section{Empirical framework}
\label{sec:framework}

We model each Arena vote as a discrete choice among the two responses and the option to reject both. A user's utility from a response depends on its accuracy, its style, and the model that produced it, and users differ in how much they value each.

Let $i$ index users, $p$ prompts, and $m$ language models. In each battle, user $i$ submits prompt $p$ and sees two responses, A and B, from two models. The user picks one response or rejects both. We index these options by $j\in\{A,B,0\}$, where $0$ is ``both bad.'' We write user $i$'s utility from model $m$'s response to prompt $p$ as $U_{ipm}=V_{ipm}+\varepsilon_{ipm}$, where $\varepsilon_{ipm}$ is an idiosyncratic shock. The systematic utility $V_{ipm}$ depends on observed characteristics and the user's tastes:
\begin{equation}
V_{ipm} = {\beta^R_{ip}}'x^R_{pm} + {\beta^M_{ip}}'x^M_m + \delta_m - \beta^A_{ip} L(s_{pm}) + \alpha\,\mathbf{1}\{\text{position B}\}.
\label{eq:utility}
\end{equation}
The superscripts $R$, $M$, and $A$ mark response features, model attributes, and accuracy. The vector $x^R_{pm}$ contains the thirteen response features and $x^M_m$ the three model attributes; $\beta^R_{ip}$ and $\beta^M_{ip}$ are user $i$'s coefficients on them. The model fixed effect $\delta_m$ captures persistent differences in utility across models that accuracy, response features, and prompt interactions do not explain. We include \EstimatedModels{} fixed effects, one for each model in the estimation sample; we drop the one battle whose model appears fewer than \MinModelBattles{} times. The coefficient $\alpha$ captures the effect of appearing on the right. The term $\beta^A_{ip}L(s_{pm})$ is the utility cost of inaccuracy, which we describe next.

The score $s_{pm}$ is the judged accuracy of model $m$'s response to prompt $p$, an integer from 0 to 10. The loss function $L(s)$ is the utility a response with score $s$ loses relative to a response with a perfect score of 10 and the same features and model, on the average prompt and for the average user. The only restriction on $L$ is that it falls with the score. We set $L(10)=0$ and
\begin{equation}
L(s)=\sum_{k=s}^{9}e^{d_k},\qquad s=0,\dots,9,
\label{eq:loss}
\end{equation}
so each step down the scale, from $k+1$ to $k$, adds a positive loss $e^{d_k}$. Holding everything else fixed, a less accurate answer is therefore never better. We estimate the ten steps $d_0,\dots,d_9$ freely. The value of accuracy can thus be nonlinear in the score, and we do not treat the judge's scale as cardinal. The coefficient $\beta^A_{ip}$ scales this loss with the prompt and the user.

The coefficients vary with the prompt, through the three prompt scores $h_p$, and across users:
\begin{equation}
\begin{aligned}
\beta^R_{ip} &= \beta^R_0 + B^R h_p + \nu^R_i, \\
\beta^M_{ip} &= \phantom{\beta^R_0 + {}} B^M h_p + \nu^M_i, \\
\beta^A_{ip} &= \beta^A_i\, e^{\theta'h_p}.
\end{aligned}
\label{eq:coefficients}
\end{equation}
The intercept $\beta^R_0$ gives mean preferences on the average prompt. We standardize the prompt scores, so $h_p=0$ is the average prompt and each column of $B^R$ and $B^M$ is the shift from a one-standard-deviation change in a score. The second equation has no intercept, because the model fixed effects absorb it. The third scales the accuracy coefficient rather than shifting it, so it stays nonnegative: $e^{\theta'h_p}$ is the prompt's multiplier, equal to one on the average prompt, and $\beta^A_i\ge0$ is the user's accuracy weight.

We assume the style and model taste deviations are normally distributed. Stacking $\nu^R_i$ and $\nu^M_i$, which have thirteen and three components, gives $\nu_i\sim N(0,\Sigma)$, where $\Sigma$ is a diagonal matrix of variances. A larger variance for a characteristic means users differ more in how much they value it.

The accuracy weight $\beta^A_i$ is nonnegative and has a censored normal distribution \citep{train2005mixed,train2009discrete}:
\begin{equation}
\beta^A_i=\max(0,\,b+\sigma^A\eta_i),\qquad \eta_i\sim N(0,1).
\label{eq:sensitivity}
\end{equation}
The parameter $\sigma^A$ governs the spread: a share $\Phi(-b/\sigma^A)$ of users put no weight on judged accuracy, and the rest weigh it more or less heavily than the average user. Only the product $\beta^A_iL(s)$ enters utility, so the level of $\beta^A_i$ and the level of $L$ cannot be told apart. We set $b$ so that the mean of $\beta^A_i$ is one, so $L$ measures the cost of inaccuracy for the average user. Appendix~\ref{app:model} compares this distribution with alternatives. We assume the taste deviations and $\eta_i$ are independent of each other, of the prompt, and of the responses' characteristics.

We write $V_{ipA}$ and $V_{ipB}$ for the systematic utilities of the two responses in a battle. The systematic utility of ``both bad'' is $V_{ip0}=\kappa'h_p$. Every prompt in Arena receives one vote, so we cannot estimate a separate value of rejecting both responses for each prompt. The term $\kappa'h_p$ lets that value vary with the prompt: on some prompts users reject both responses more readily, whatever the responses. We assume the shocks $\varepsilon_{ipj}$ are i.i.d.\ standard type-I extreme value, independent of the tastes. These assumptions fix the location and scale of utility, so we measure utility in logits.

We integrate the logit probability over the taste distribution to obtain each choice probability:
\begin{equation}
P_{ip}(j)=\int \frac{\exp\{V_{ipj}(\nu,\eta)\}}{\exp\{V_{ip0}\}+\exp\{V_{ipA}(\nu,\eta)\}+\exp\{V_{ipB}(\nu,\eta)\}}\,dF(\nu,\eta),
\label{eq:choice-probability}
\end{equation}
where $F$ is the joint distribution of the tastes. Given a user's tastes, and conditional on not rejecting both responses, the user picks response A over B with probability $\sigmoid(V_{ipA}-V_{ipB})$, where $\sigmoid(x)=1/(1+e^{-x})$ is the logistic function. We use this probability to express utility differences as win rates.

When the user declares a tie, the battle contributes $\tfrac12\log P_{ip}(A)$ plus $\tfrac12\log P_{ip}(B)$ to the objective, as if it were half a choice of each response. This term is not the probability of a tie, so the objective is a pseudo-likelihood. A ``both bad'' vote is the choice of the outside option.

We treat battles as independent, so we estimate the population distribution of tastes, not the tastes of any particular user. We identify the taste distribution from how the probabilities of picking each response and of rejecting both change with the two responses' characteristics across battles, as in random-coefficients demand estimation with cross-sectional data \citep{berry1995automobile,nevo2001measuring,berry2014identification}. Arena assigns models to battles at random, so the models a user sees are independent of the user's tastes. Voting is voluntary, however, and a user may be more likely to vote when a response suits their tastes. We therefore assume that the taste distribution is the same across the model pairs and responses in the voted sample. Independence from the prompt, which the user writes, and from the responses' characteristics is also an assumption. Because we treat battles as independent, the spread of tastes may also reflect variation in a user's tastes from one battle to the next (Section~\ref{sec:conclusion}).

We estimate the parameters by simulated maximum likelihood \citep{mcfadden2000mixed}.\footnote{The model has \ModelParameters{} parameters. We approximate the integral with \QuasiRandomDraws{} quasi-random draws.} In simulated data, our estimator recovers the mean coefficients, fixed effects, accuracy parameters, and large standard deviations closely; smaller standard deviations can be biased or come out near zero (Appendix~\ref{app:model}).

We compute standard errors from the inverse Hessian of the objective, treating battles as independent and the prompt scores as fixed. Because ties enter with half weights and some battles share a voter, these standard errors are approximate. They also omit uncertainty from estimating the prompt projection, which uses the same training battles. Appendix~\ref{app:data} reports a split-sample check of the projection's stability.

\section{Preference estimates}
\label{sec:estimates}

\begin{table}[!tbp]
\caption{Estimates of the preference model.}
\label{tab:model}
\centering
\footnotesize
\input{tables/tab_estimates.tex}
\par\vspace{4pt}
\begin{minipage}[t]{\linewidth}
\scriptsize\textit{Note:} Panels A and B report, for each coefficient in Equation~\eqref{eq:coefficients}, its mean on the average prompt ($h_p=0$) and its standard deviation across users, in logits. Panel C reports the loss $L(s)$ in Equation~\eqref{eq:loss} for the average user on the average prompt, in logits; the scale $\sigma^A$ of the latent normal behind the accuracy weight $\beta^A_i$ in Equation~\eqref{eq:sensitivity}; and the share of users with $\beta^A_i=0$. We choose $b$ so that $\beta^A_i$ has mean one; its standard deviation is \AccuracyWeightSD. We standardize length, Flesch reading ease, and the list, header, and bold-span counts; the other response features are indicators. Standard errors are in parentheses.
\end{minipage}
\end{table}

\begin{figure}[t]
\centering
\includegraphics[width=0.88\linewidth]{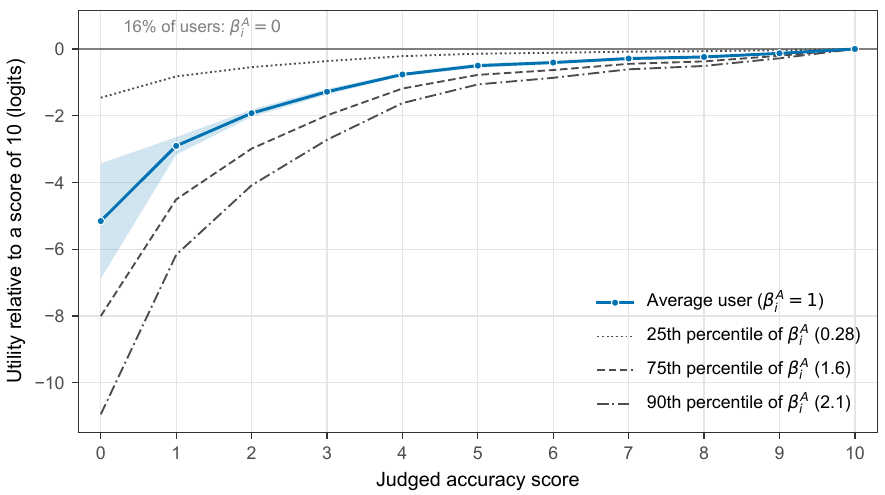}
\caption{The utility of a response with judged score $s$ relative to a response scored 10 with the same features and model, on the average prompt, $-\beta^A_iL(s)$. The solid line is the average user ($\beta^A_i=1$), with a 95\% confidence band; the other lines are users at the 25th, 75th, and 90th percentiles of the accuracy weight $\beta^A_i$. Users with $\beta^A_i=0$ lie on the horizontal axis.}
\label{fig:accuracy}
\end{figure}

Table~\ref{tab:model} reports the estimates of the preference model in Equations~\eqref{eq:utility}--\eqref{eq:sensitivity}, and Figure~\ref{fig:accuracy} plots the cost of inaccuracy; Appendix~\ref{app:model} reports the prompt interactions and the model fixed effects. Judged accuracy has the largest effect on choice, and most of its value lies at the bottom of the scale. Conditional on not rejecting both responses, a user with mean tastes picks a response with a perfect score over a response scored 7 with the same features and model only \PerfectOverSeven{} of the time. The same user picks a response scored 7 out of 10 over one scored 3 out of 10 \SevenOverThree{} of the time and almost never picks a response scored 0. Few responses score 0, so the loss at that score is imprecise. By comparison, a response one standard deviation longer wins \LongerWin{} of the time.

The value of accuracy varies across prompts and users. The first prompt score is highest for prompts whose winning answers are long and heavily formatted, with lists, bold text, tables, and links. A one-standard-deviation increase in this score multiplies the cost of inaccuracy by \AccuracyMultiplier{}, which lowers the win rate of a perfect response over one scored 7 from \PerfectOverSeven{} to \PromptPerfectOverSeven{}. Across users, \ZeroAccuracyWeight{} put no weight on judged accuracy, while the user at the 75th percentile weighs it \AccuracyUpperQuartile{} times as much as the average user. The zero share rests on our restriction to nonnegative weights: without it, the data fit as well with \NegativeAccuracyWeight{} of users who mildly prefer less accurate answers (Appendix~\ref{app:model}). Part of this variation may reflect judge error on answers that are hard to grade: a second judge agrees with the first on most high scores but often disagrees on intermediate ones (Appendix~\ref{app:accuracy}).

Prompts also affect users' willingness to reject both responses. Raising the first prompt score by one standard deviation raises the value of ``both bad'' by \RejectBothRise{} logits.

Holding judged accuracy fixed, users differ most in their preferences for links, emoji, and tables, followed by hedging, length, and bold spans (Table~\ref{tab:model}). These standard deviations are per unit of each feature: the presence of a link, emoji, or table, or one standard deviation of length or bold spans. A feature that is rare, such as a link, can therefore have a large standard deviation yet contribute less to the spread of utility across battles. For each of these features except length, the standard deviation across users exceeds the absolute value of the mean, so many users have the opposite taste from the average user. For a user with mean tastes on the average prompt, a response with a table wins \TableWin{} of the time against an otherwise identical one without, yet \DislikeTables{} of users prefer the response without a table. Similarly, \DislikeEmoji{} of users dislike emoji, and \PreferShorter{} prefer shorter responses. For positive and negative tone, the estimated standard deviations are near zero.

Preferences for length also vary with the prompt. A one-standard-deviation increase in the second prompt score raises the win rate of a response one standard deviation longer from \LongerWin{} to \LongerPromptWin{}. By contrast, prompt scores move the value of tables far less than differences in tastes across users do. Appendix Table~\ref{tab:interactions} reports the interactions.

Models with higher benchmark scores have larger fixed effects. The fixed effects capture persistent differences in utility across models that judged accuracy and observed style do not explain, such as capability and default behavior (Appendix Figure~\ref{fig:modeleffects} plots the \EstimatedModels{} effects against the benchmark index). We regress the \EstimatedModels{} fixed effects on the model attributes. Ten more points on the benchmark index are associated with a fixed effect \BenchmarkCoef{} logits higher ($t=\BenchmarkT{}$): this fitted difference alone would make a response win \BenchmarkWin{} of the time against an otherwise identical one. It is not a prediction for two particular models ten points apart, which also differ in their other attributes and in the part of their fixed effects the index does not explain. We read this coefficient as a description, not a causal effect. Models with higher benchmark scores also differ in ways we do not observe, for example in how much post-training they receive for conversation. When we run the same regression on fixed effects from standard logits, the benchmark coefficient falls by about half once response characteristics enter (Appendix~\ref{app:model}). This decline is consistent with more capable models earning part of their premium through more accurate and better-formatted responses.

\section{Routing and steering}
\label{sec:results}

We use the estimated preferences to evaluate how much personalization is worth and how much of that value a platform can capture.

\subsection{A framework for routing and steering}
\label{sec:routing-framework}

A platform can choose which model answers a request and what instruction, if any, to add to the prompt. Write $V_{ipm}(a)$ for user $i$'s systematic utility from model $m$'s answer to prompt $p$ with instruction $a$. With no added instruction, $V_{ipm}(0)=V_{ipm}$ from Equation~\eqref{eq:utility}. Let $\mathcal A$ be the available pairs of models and instructions and $\mathcal I$ the information available when choosing among them. The largest expected gain over always serving the best fixed model without instructions is
\begin{equation}
G(\mathcal I,\mathcal A)
= \mathbb E\!\left[\max_{(m,a)\in\mathcal A} \mathbb E[V_{ipm}(a)\mid\mathcal I]\right]
  - \max_m \mathbb E[V_{ipm}(0)].
\label{eq:routing-gain}
\end{equation}
Routing alone varies $m$ with $a=0$. Steering alone holds the model fixed and varies $a$. Suppose the platform has just one alternative to the baseline, either switching to another model or asking the baseline model for a shorter answer. If $d(\mathcal I)$ is the expected utility change from taking that action, the gain is $\mathbb E[\max\{d(\mathcal I),0\}]$. For the instruction, this calculation weighs the value of brevity to the user against any loss in accuracy from omitting an important detail. With both routing and steering, the platform can choose the model it expects to give the best brief answer when asked.

Two kinds of information enter this comparison. On the preference side, past choices can reveal persistent tastes, and the current prompt can reveal a preference specific to this request. Our simulations assume that user tastes $\nu_i$ and $\beta^A_i$ are fixed across prompts. On the response side, before generation the platform must forecast each model's response characteristics and how an instruction would change them; only differences in expected utility across the available options affect its choice. After generation, the platform observes each answer's realized style and accuracy, including variation across draws that even a perfect forecast of mean characteristics would miss.

Our four exercises vary $\mathcal I$ and $\mathcal A$ in turn. Section~\ref{sec:routing} holds $a=0$ and includes all realized responses in $\mathcal I$, comparing prompt-only information with full knowledge of the user's tastes. Section~\ref{sec:learning} replaces known tastes in $\mathcal I$ with a history of pairwise choices. Section~\ref{sec:exante} removes realized responses from $\mathcal I$, forcing the platform to choose a model before generation from the prompt, the user's choice history, or both. Section~\ref{sec:steering} expands $\mathcal A$ with instructions built from the user's tastes before generation, first holding the model fixed and then letting a combined policy choose among the realized steered and unsteered answers across models.\footnote{
  Separately from the expected utility to the user from the response, routers also differ in their costs and latencies. We are not able to measure these factors in our study.}

\paragraph{Panel and evaluation.} We use the panel described in Section~\ref{sec:data}: \PanelPrompts{} prompts with one response from each of ten models, GPT-4.1, GPT-4o, Claude Sonnet 4, Gemini 2.5 Pro, DeepSeek V3, Llama 4 Maverick, Grok 4, Qwen3 235B, Kimi K2, and MiniMax M1. We simulate users by drawing taste vectors $\nu_i\sim N(0,\Sigma)$ and accuracy weights $\beta^A_i$ from the estimated distributions, and evaluate each response with the user's systematic utility in Equation~\eqref{eq:utility}. Our baseline serves every user and prompt the same model without instructions, Gemini 2.5 Pro, which has the highest average utility in the panel. It is the model a platform that does not personalize would choose. The routing and learning results are counterfactuals: we use the estimated preferences to predict how users would fare under policies that Arena does not use. The steering results use newly generated responses, but we also score them with the estimated preferences. We collect no new human votes.

For policy $r$, let $(m,a)=r(p,i)\in\mathcal A$ denote the model and instruction it chooses for user $i$ on prompt $p$. No policy observes the logit shocks, so the expected utility gain over the baseline on that request equals the difference in systematic utility, $\Delta V_{ipr}=V_{ipm}(a)-V_{ip,\text{Gemini}}(0)$. Averaging $\Delta V_{ipr}$ over users and prompts gives the policy's mean gain, which attains $G(\mathcal I,\mathcal A)$ in Equation~\eqref{eq:routing-gain} when the policy maximizes expected utility given $\mathcal I$. We report both the mean gain and the win rate. For each user and prompt, $\sigmoid(\Delta V_{ipr})$ is the probability that the user picks the policy's response over Gemini's, conditional on not rejecting both (Section~\ref{sec:framework}). The win rate is the average of these probabilities over users and prompts. It integrates over the shocks of the two responses. Because gains vary across users and prompts, the win rate differs from $\sigmoid$ of the mean gain. Where a result requires simulated choices, as when a platform learns a user's tastes, we draw each choice from the logit choice probabilities, so the shocks make the simulated choices noisy. The intervals we report describe variation across simulated users, conditional on the estimated preferences and the response panel. They omit uncertainty in the estimated preferences. When we redraw the preferences from their estimated sampling distribution, steering Gemini 2.5 Pro still beats routing across ten models, and the user-by-prompt router still beats the prompt router, in nearly every draw; the smaller gap between one model per user and the prompt router in Section~\ref{sec:exante} is not reliably signed.

\subsection{The value of personalization}
\label{sec:routing}

We first evaluate the full-information routing benchmarks in Equation~\eqref{eq:routing-gain}, holding $a=0$ and letting the platform observe all ten realized responses. For each prompt, the prompt router knows only the prompt and the responses; because utility is linear in tastes, expected utility given this information is the average user's utility $V_{0pm}$, so it serves $\arg\max_m V_{0pm}$. The user-by-prompt router also knows the user's tastes and serves $\arg\max_m V_{ipm}$. Neither router observes the logit shocks, so both attain $G(\mathcal I,\mathcal A)$ for their respective information sets. We evaluate both on \SimulatedUsers{} simulated users.

\begin{table}[t]
\caption{Utility gains and win rates of the two routers. The baseline always serves Gemini 2.5 Pro. Gains are in logits. The win rate is the probability that the user prefers the first policy's response to the second's, averaged over \SimulatedUsers{} simulated users and the prompts. Pairs where both policies serve the same response count as half a win. The last three columns use only the user-prompt pairs where the two policies serve different responses; ``Share'' is their fraction of all pairs. Panel B holds the number of generated responses at ten but draws all ten from Gemini 2.5 Pro, on the \BestOfNPrompts{} prompts for which it answered nine more times (Appendix~\ref{app:bestofn}).}
\label{tab:routing}
\centering
\small
\setlength{\tabcolsep}{3.5pt}
\input{tables/tab_routing.tex}
\end{table}

The prompt router wins against the baseline \PromptWin{} of the time (Table~\ref{tab:routing}). Its gain is small because it serves the baseline model itself on \PromptKeepsBaseline{} of prompts, where its gain is zero. The user-by-prompt router wins \UserPromptWin{} of the time. It can always pick the prompt router's choice, so it never does worse, and it picks a better response in \PersonalImproves{} of user-prompt pairs. When it serves a response other than the baseline, which it does for \OracleSwitches{} of user-prompt pairs, the user prefers that response \OracleSwitchWin{} of the time, a gain of \OracleSwitchGain{} logits. Knowing the user therefore accounts for \PersonalGainShare{} of the gain in mean utility, \PersonalGain{} of \OracleGain{} logits. This share is the value of adding the user's tastes to a router that already sees the prompt and every response, and it depends on how much of the prompt's effect our three prompt scores capture. Other orderings give other shares: knowing only the user's tastes and serving one model per user, without seeing the responses, gains \KnownUserGainShare{} as much as the user-by-prompt router (Section~\ref{sec:exante}).

Part of these gains may come from picking a lucky draw rather than a better model: the same model gives a different response each time it answers a prompt (Section~\ref{sec:exante}). To separate the two, we hold the number of responses fixed. We draw nine more responses from Gemini 2.5 Pro for each panel prompt and let each router pick among ten Gemini responses instead of ten models (Table~\ref{tab:routing}, Panel B; Appendix~\ref{app:bestofn} varies the number of responses). This comparison holds the number of responses fixed, not their cost. It also does not separate a model's persistent fit to a user from differences in how much each menu's responses vary from draw to draw. For the user-by-prompt router, the best of ten Gemini responses gains \DrawsPersonalGain{} logits over the baseline, about two thirds of the \TenModelPersonalGain{} from ten models. For the prompt router, ten Gemini responses do slightly better than ten models, \DrawsPromptGain{} against \TenModelPromptGain{} logits.

First, extra draws are worth more when the router knows the user: from the same ten Gemini responses, the user-by-prompt router gains \DrawsPersonalGain{} logits and the prompt router \DrawsPromptGain{}. Second, a menu of different models beats a menu of Gemini draws only when the router knows the user. The prompt router's pick from ten models loses to its pick from ten Gemini responses, winning \ModelsBeatDrawsPrompt{} of the time. The user-by-prompt router's pick from ten models wins \ModelsBeatDrawsPersonal{} of the time and adds \ModelsOverDrawsGain{} logits, more than the prompt router's entire gain. Users differ in which model suits them, so a menu of models is valuable only to a router that knows them. Third, personalization does not need extra draws when the platform knows the user's tastes: with one response per prompt, serving each user one model chosen by their tastes gains about as much as the prompt router with all ten responses in hand. With tastes learned from \FeedbackRounds{} choices, the same rule gains less than the prompt router (Section~\ref{sec:exante}).

Users disagree about which model is best. Under the user-by-prompt router, the leading model serves \LargestShare{} of user-prompt pairs, and the eight models outside the top two together serve \OtherModelsShare{} (Table~\ref{tab:shares}). Even if each user had to choose one model for all prompts, the best model for \PersonalBestDiffers{} of users would differ from the best model for the average user. Both kinds of taste contribute. With only style tastes, the best model differs for \BestDiffersStyleOnly{} of users; with only tastes for model attributes, which shift a model's utility on every prompt, it differs for \BestDiffersModelOnly{}. Averaging over user-prompt pairs, the models selected by the user-by-prompt router score \OracleBenchmark{} on the Artificial Analysis index, compared with \FixedBaselineBenchmark{} for always serving Gemini 2.5 Pro and \PromptBenchmark{} for the prompt router.

\begin{table}[t]
\caption{Assignment shares of the user-by-prompt router over user-prompt pairs. Users may switch models across prompts.}
\label{tab:shares}
\centering
\small
\input{tables/tab_shares.tex}
\end{table}

\subsection{Learning tastes from pairwise feedback}
\label{sec:learning}

The user-by-prompt router knows each user's tastes; a platform must learn them. We simulate a platform that knows the estimated population parameters but not the user's taste deviation $\nu_i$. In each round, the user submits a new prompt, and the platform shows two responses: one from the model it currently estimates is best for the user, and one from a randomly chosen other model. The user picks one or rejects both, with the logit probabilities under the user's true tastes. After each round, the platform updates its beliefs about the user's tastes by Bayes' rule, using the estimated taste distribution as the prior. Let $y_{ir}\in\{A,B,0\}$ be the user's choice in round $r$. The platform also learns the user's accuracy weight. It replaces the distribution of $\beta^A_i$ with a finite set of types $q$, each with weight $a_q$ and prior share $w_q$ (Appendix~\ref{app:learning}). Given $\nu_i$, the type, and the responses shown in each round, each choice follows the logit probability, and the logit shocks are independent across rounds. The posterior density of the user's tastes and type after $t$ rounds is
\begin{equation}
p(\nu,q\mid y_{i1},\dots,y_{it})\ \propto\ w_q\,\phi(\nu;\Sigma)\prod_{r=1}^{t}\ell_{ir}(\nu,a_q),
\label{eq:posterior-density}
\end{equation}
where $\ell_{ir}(\nu,a)$ is the logit probability of $y_{ir}$ given tastes $\nu$ and accuracy weight $\beta^A_i=a$, the fraction inside the integral in Equation~\eqref{eq:choice-probability}. For each type, the platform computes the posterior mode of the tastes. Taking logs and dropping constants,
\begin{equation}
\hat\nu_{itq}=\arg\max_{\nu}\ \sum_{r=1}^{t}\log \ell_{ir}(\nu,a_q)-\tfrac{1}{2}\,\nu'\Sigma^{-1}\nu.
\label{eq:posterior}
\end{equation}
Because utility is linear in tastes, maximizing expected utility given the choice history means routing at the posterior mean tastes. We use a Laplace approximation within each accuracy type to obtain the posterior probabilities of the types, $\hat\pi_{itq}$, and route using approximate posterior mean tastes, $\hat\nu_{it}=\sum_q\hat\pi_{itq}\hat\nu_{itq}$ and $\hat\beta^A_{it}=\sum_q\hat\pi_{itq}a_q$. Before any feedback, $\hat\nu_{i0}=0$ and the platform serves the prompt router's choice. In later rounds, and on held-out prompts, it serves the model with the highest systematic utility at $\hat\nu_{it}$ and $\hat\beta^A_{it}$. We simulate \LearningUsers{} users, drawing their tastes from the estimated distributions, and evaluate the platform's choices on \LearningEvalPrompts{} prompts not used for feedback (Appendix~\ref{app:learning}).

\begin{figure}[t]
\centering
\includegraphics[width=0.88\linewidth]{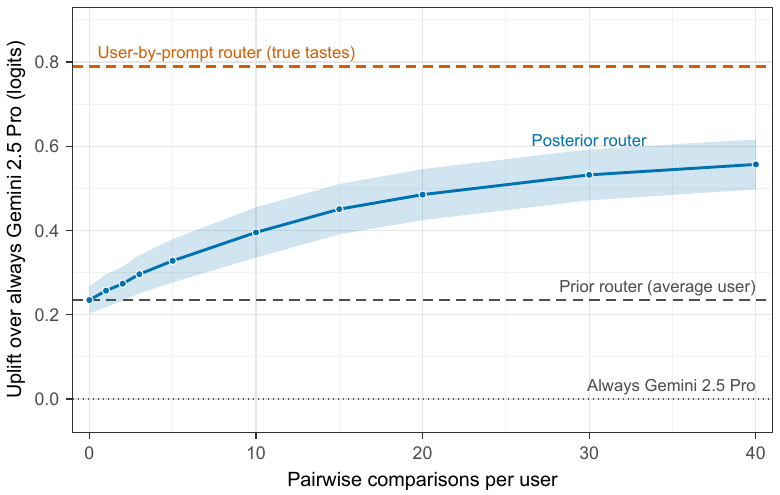}
\caption{Learning tastes from pairwise feedback, for \LearningUsers{} simulated users on \LearningEvalPrompts{} held-out prompts. The line is the learning router's mean utility gain over the baseline (always Gemini 2.5 Pro), in logits, by the number of pairwise comparisons per user; the band is a 95\% interval over users. Dashed lines mark the prompt router, which uses the average user's tastes, and the user-by-prompt router, which knows each user's tastes.}
\label{fig:learning}
\end{figure}

Before any feedback (Figure~\ref{fig:learning}), the router serves the average user's best response and wins against the baseline \PriorWin{} of the time. After 10 comparisons it closes \TenChoicesClosed{} of the gap to the user-by-prompt router. After \FeedbackRounds{} it closes \FortyChoicesClosed{} and wins \LearnedWin{} of the time. A short history of choices therefore recovers most of the value of knowing a user's tastes, provided that those tastes persist across prompts.

\subsection{Choosing a model before generation}
\label{sec:exante}
\label{sec:noise}

The routers above require a response from every model, so the platform generates ten responses to serve one. A platform that wants to avoid this cost must choose a model before any response exists. At that point it can observe the prompt and the user's past choices. We ask which of the two is more valuable.

Each rule we study predicts the utility of every model's response before it is generated and serves the model with the highest prediction (Appendix~\ref{app:exante}). A prompt-only rule predicts each model's response characteristics, its judged-accuracy loss $L(s)$ and its thirteen response features, from the prompt's embedding, using regressions fit on other prompts. It evaluates the average user's utility at the predicted characteristics, so it serves the same model to every user who submits the prompt. Predicting utility directly, or using a neural network, does no better.

A user-only rule learns the user's tastes from $k$ past choices. Arena pairs models at random, so we give each user a history of $k$ choices between randomly paired models and estimate the user's tastes as in Section~\ref{sec:learning}. The rule serves the model with the highest utility at the estimated tastes, averaged over prompts, so each user gets one model for all prompts. As a benchmark, we also apply the rule with the user's true tastes. A combined rule evaluates the user's tastes at the characteristics predicted from the prompt, so its choice varies across both users and prompts.

We reserve \ExanteLearnShare{} of the prompts for users' choice histories and evaluate every rule on the other \ExanteEvalShare. We cross-fit the prompt-side predictions over all prompts, so no prompt's own responses inform its prediction. We score each rule by the user's true utility from the response that the chosen model actually produced, and compare it with the two routers of Section~\ref{sec:routing}, which observe every response, on the same sample.

\begin{table}[t]
\caption{Choosing a model before generation, on held-out prompts and users. Gains are over the baseline (always Gemini 2.5 Pro), with 95\% intervals over users; the win rate is against the baseline. Panel A repeats the routers of Table~\ref{tab:routing} on this sample. Rules in Panels B and D learn tastes from a user's past choices between randomly paired models; ``known tastes'' uses the user's true tastes. Panel C reports the best of the prompt-only rules, which predicts each model's response characteristics from the prompt; Appendix Table~\ref{tab:exante-full} reports the other rules.}
\label{tab:exante}
\centering
\small
\input{tables/tab_exante.tex}
\end{table}

Our prompt-only rules capture little of the gain available after generation (Table~\ref{tab:exante}). The best of them wins against the baseline \ExantePromptWin{} of the time and gains \PromptCeilingShare{} as much as the prompt router that observes the responses. A platform that knows a user's tastes can assign one model per user that gains \KnownUserGainShare{} as much as the user-by-prompt router, about as much as the prompt router gains with all ten responses in hand. A taste learned from \FeedbackRounds{} past choices recovers three fifths of that gain. The two kinds of information are complements: the combined rule with a learned taste wins against the baseline \ExanteCombinedWin{} of the time, the highest among the rules we evaluate that a platform could implement. It still falls \CombinedGap{} logits short of the user-by-prompt router. Learning tastes rather than knowing them accounts for \TasteGap{} logits of this gap; the other \ResponseGap{} is the value of observing the responses rather than relying on our predictions of them. This split depends on the order of the two steps: we first replace learned tastes with known ones and then predicted responses with realized ones. The second step combines the value of picking a favorable draw with the shortfall of our forecasts of each model's expected utility. On the held-out routing sample, our fitted forecasts capture \FittedForecastUserShare{} of the user-by-prompt router's gain and \FittedForecastPromptShare{} of the prompt router's. On the prompts in the sampling audit, a rule that instead uses each model's mean utility over four draws for the prompt captures \ForecastCeilingUserShare{} and \ForecastCeilingPromptShare{}. Observing the realized response is thus worth about \SeeDrawUserShare{} of the personalized gain, and much of the shortfall of the prompt-only rules comes from our forecasts rather than from sampling noise.

Sampling noise helps explain why observing the responses before choosing is worth so much. It does not tell us how much better predictors could do before generation. The same model gives a different response each time it answers the same prompt. To measure this noise, we have nine of the ten models answer \AuditPrompts{} prompts \AuditDraws{} times each (Appendix~\ref{app:anova}). Holding the model and prompt fixed, and scored at the average user's tastes, the utility of a response has a standard deviation of \DrawSD{} logits, compared with \InteractionSD{} logits for a model's prompt-specific advantage, the variation a prompt router exploits. One draw therefore measures that advantage with a reliability of only \DrawReliability{}, which limits how much of a single response's utility any rule based on the prompt can predict. This bound does not tell us how much a better rule could gain before generation. Our prompt-only rules also leave much of the persistent part unexplained. Tastes, by contrast, act on model characteristics averaged over many prompts, and averaging removes most of the noise.

\subsection{Personalizing one model with language}
\label{sec:steering}

Routing after generation requires access to many models and a response from each. A platform can instead tell one model what the user likes. We turn each of \SteeringProfiles{} simulated users' strongest style tastes into at most \MaxInstructions{} plain-language instructions, such as a preference for concise answers or for tables (Appendix~\ref{app:verbal}). Gemini 2.5 Pro answers \SteeringPrompts{} prompts once without instructions and once for each user. We judge every response for accuracy and score it with the user's utility.

\begin{table}[t]
\caption{Steering Gemini 2.5 Pro and comparison with routing. Panel A compares a steered response with a fresh unsteered response from Gemini 2.5 Pro to the same prompt, evaluated under different users' tastes (\SteeringProfiles{} profiles $\times$ \SteeringPrompts{} prompts). Panel B compares each policy with always serving Gemini 2.5 Pro's unsteered panel response, under the specific user's tastes, on the \MatchedSteeringPairs{} user-prompt pairs in which every steered and fresh unsteered response exists. The ``two unsteered'' row routes over the panel and fresh unsteered responses of the six models we steer; the next row chooses one of these models before generation and serves its steered response; the last row routes over their unsteered and steered responses. Win rate is the mean over user-prompt pairs of the predicted probability that the user prefers the policy's response.}
\label{tab:verbal}
\centering
\small
\input{tables/tab_verbal.tex}
\end{table}

Our estimates predict that steering benefits the intended user but, on average, not other users. A response steered toward a user wins against a fresh unsteered response \verbWin{}\% of the time under that user's tastes, but only \SteeringAverageWin{} under the average user's tastes. Under another user's tastes, its mean gain is slightly negative (Table~\ref{tab:verbal}, Panel A). We attribute nearly all of the utility gain to links, emoji, tables, and length; changes in judged accuracy contribute little.

Steering Gemini 2.5 Pro performs about as well as routing across ten models. Against the baseline, it wins \SteeredPanelWin{} of the time, compared with \SteeringSubsetOracleWin{} for the user-by-prompt router over all ten models on the same users and prompts (Panel B). The two approaches combine well: routing over the six steered models' original and steered responses reaches \CombinedSteeringWin{}. Most of this gain does not come from a larger menu: routing over two unsteered responses from each of these models wins \TwoUnsteeredWin{} of the time. With one response per request, choosing a model from the user's tastes before generation and then steering it wins \ChooseThenSteerWin{}.

The comparison depends on links. Links account for \verbLinkSh{}\% of Gemini 2.5 Pro's steering gain over a fresh unsteered response, more than any other feature. Steering and routing yield gains of similar size, and which is larger depends on how users value links. If links were worth half as much, steering would win \LinkHalfSteerWin{} of the time and routing \LinkHalfRouterWin{}; if links were worth nothing, steering would win \LinkZeroSteerWin{} and routing \LinkZeroRouterWin{}. Combining the two would still win \LinkZeroCombinedWin{}.

Steering works for every model we test. For Gemini 2.5 Pro, with tastes learned from \FeedbackRounds{} choices, steering recovers \LearnedSteeringShare{} of the gain from known tastes (Appendix~\ref{app:verbal}).

\section{Conclusion}
\label{sec:conclusion}

Users disagree about what makes a language model's response good. We estimate these disagreements from Arena votes and find that they are large enough to change which model is best for a user and a prompt. We then measure how much a platform could gain by personalizing its responses.

Most of the value of personalization comes from knowing the user. Against always serving the model with the highest average utility, routing each prompt to the best model for the average user wins \PromptWin{} of the time, and routing on the user's tastes as well wins \UserPromptWin{}. A platform can learn much of this value quickly: after \FeedbackRounds{} pairwise comparisons, it wins \LearnedWin{} of the time. Without generating responses from every model, a platform that combines a learned taste with characteristics predicted from the prompt wins \ExanteCombinedWin{} of the time. Telling a single model what the user likes wins \SteeredPanelWin{} of the time, about as often as routing across ten models, although this comparison depends on how users value links.

These results bear on competition among model providers. A router gives differentiated models demand on the prompts and users for which they have a comparative advantage, and it spares users the work of choosing a model. If users instead stay with a single provider, the industry may concentrate on the model with the highest average utility and underserve less common tastes. Routing power may also concentrate in an intermediary whose objectives, fees, or default rankings shape traffic.

Our analysis has several limitations. First, we score every policy with the estimated preference model, and the accuracy of our results depends on how well the choice model captures true user preferences.\footnote{For example, the choice model values a link whether or not it works, and some of the links that steering adds may not work.} Some of the randomness we model as logit shocks may reflect systematic preferences or response features that we do not model. Relatedly, Arena links only some battles to the same voter, mostly within one sitting, and our estimate does not use these links. We therefore cannot separate persistent differences in tastes across users from changes in a user's tastes from one battle to the next. All our routing, learning, and steering results assume that these taste differences persist across a user's prompts. The value of learning from a user's history depends on how much of that variation persists. Repeated choices by the same users over a longer period would separate the two. Second, our results abstract from latency, cost, and users' reactions to being routed.
Third, Arena users are self-selected and anonymous, and their tastes may differ from those of other users. Field experiments with AI platforms could address these limitations.

Finally, because Arena users pay nothing, we measure utility in logits rather than dollars. Estimates of willingness to pay would let a platform compare routing gains directly to inference costs. In practice, however, if most consumers use free, advertising-supported services rather than paid subscriptions, platforms will weigh inference costs against engagement and ad revenue instead.

\clearpage

\input{paper_full.bbl}
\clearpage
\appendix

\section{Data construction}
\label{app:data}

\begin{table}[h]
\caption{Sample construction and outcomes. The all-language count is the documented pre-filter count from the two source releases; all other counts are computed from the processed data.}
\label{tab:sample}
\centering
\small
\input{tables/tab_sample.tex}
\end{table}

We use the Arena \texttt{100k} and \texttt{140k} human-preference releases on Hugging Face. \EmptyResponseBattles{} battles in which either response is empty are dropped from the corpus before every count in Table~\ref{tab:sample} except the all-language count, because an empty response is a failed generation rather than a style. The English corpus is \OldReleaseShare{} from the 2024 release and \NewReleaseShare{} from the 2025 release. We remove \IncompleteBattles{} battles that lack complete features or a usable outcome, and one battle with a model that appears fewer than \MinModelBattles{} times. We embed prompts with OpenAI's \nolinkurl{text-embedding-3-large}. The PLS projection predicts the differences between winners and losers in fourteen response features: the features in Section~\ref{sec:data} and the SMOG readability index, with length in logs. We divide each difference by its standard deviation, so that no feature dominates the projection through its units. It uses only training battles with a winner; ties and ``both bad'' votes have no winner and do not enter the projection. We leave SMOG out of the choice model because its rank correlation with Flesch reading ease is \SmogFleschCorr. We estimate the standardization moments of the continuous features and the PLS projection of the prompt embeddings on the training split only, and apply them to held-out battles and to all simulation prompts and responses.

The standard errors in the main text treat the prompt scores as known. To check the stability of the projection, we fit it separately on two random halves of the training battles. The first score, which drives almost all of the variation across prompts in the value of accuracy, correlates at \PlsSplitCorr{} across the two fits.

We classify sentiment and hedging with \nolinkurl{openai/gpt-oss-20b}, called through OpenRouter at temperature zero. The same classifier assigns prompts to seven topics. It labels the prompts of \ClassifiedPrompts{} of the \EnglishBattles{} English battles and returns no topic for the other \UnclassifiedPrompts{}. Of the \ClassifiedPrompts{} classified prompts, \TopicInformation{} are information seeking, \TopicWriting{} writing, \TopicGuidance{} practical guidance, \TopicCoding{} coding, \TopicMath{} mathematics, \TopicConversation{} casual conversation, and \TopicMultimedia{} multimedia.

\section{Accuracy judge}
\label{app:accuracy}

The accuracy judge, \nolinkurl{z-ai/glm-5.2} called through OpenRouter at temperature zero, sees the full prompt and one response and returns a score from 0 to 10 with a short rationale. We use the same judge and rubric for the estimation sample, the ten-model panel, the sampling audit, and the steered responses. The rubric asks the judge to score only factual accuracy and correctness, not style, formatting, length, tone, or helpfulness. A score of 10 means that every verifiable claim is correct and any code would run; 6 or 7 allows a minor error or an unsupported claim; 0 or 1 means the response is fundamentally wrong, fabricated, or fails to address a factual question. For purely creative or subjective prompts with no verifiable content, the judge scores internal consistency and whether the response follows the instructions. The rubric does not separately address responses with no checkable claims. Across the \EnglishScoredResponses{} responses in the English corpus, before we drop battles with incomplete features, the scores have a mean of \ScoreMean{} and a standard deviation of \ScoreSD. On the ten-model panel, \PanelLowShare, \PanelMidShare, and \PanelHighShare{} of responses fall in the low, mid, and high bins.

The judge's scores predict votes. Among the \ScorePairs{} battles with a winner or a tie (counted as half a win) in which the two responses receive different scores, the higher-scored response wins \HigherScoreWins{} of the time. Among the \BinPairs{} of these battles whose responses fall in different bins, the higher-bin response wins \HigherBinWins{} of the time.

To validate the judge, a second judge, \nolinkurl{anthropic/claude-opus-4.8}, scores both responses in \ValidationBattles{} battles. It sees the full prompt and response but not the first judge's score, the model identities, or the vote. \ValidationRandomBattles{} battles are a random draw from an earlier \EarlierJudgeSample-battle validation sample. The other \ValidationOversampledBattles{} oversample battles in which the first judge scored either response between 3 and 7, where bins are hardest to assign, spread evenly across releases and prompt topics. The second judge returns usable scores for \ValidationScored{} of the \ValidationResponses{} responses, and for both responses in \ValidationCompleteBattles{} battles. We weight the sample to represent the full corpus: each battle's weight is the number of corpus battles in its stratum, release, and topic cell divided by the number sampled from that cell, and both of its responses carry that weight. Brackets give 95\% battle-bootstrap intervals.

The two judges place a response in the same bin \JudgeBinAgreement{} of the time (Cohen's kappa \JudgeKappa). The second judge keeps \HighBinKept{} of high responses in the high bin but only \MidBinKept{} of mid responses in the mid bin: \MidBinDown{} move to low and \MidBinUp{} to high. It moves \LowBinUp{} of low responses up. Within a battle, the judges order the two responses' scores the same way, ties included, in \JudgeScoreOrderAll{} \JudgeScoreOrderAllCI{} of battles, and in \JudgeScoreOrderDecisive{} \JudgeScoreOrderDecisiveCI{} of the battles in which both judges separate the two responses. Comparing bins instead of scores, these rates are \JudgeBinOrderAll{} and \JudgeBinOrderDecisive{}. The weighted mean absolute difference between their scores is \JudgeAbsDiff{} points; in the unweighted sample, the scores correlate at \JudgePearson{} (Spearman \JudgeSpearman). Disagreement is not concentrated in the style features whose tastes we estimate: bin mismatch is not significantly related to length, tables, or emoji at the 5\% level. It is concentrated in battles with prompts longer than \JudgePromptChars{} characters or responses longer than \JudgeResponseChars{} (\LongMismatch{} against \ShortMismatch).

To make the bins concrete, we give one example of each from the earlier validation sample. Responses are quoted in part, and the rationale is the judge's own.
\begin{itemize}
\item \textbf{Low (score 0).} Prompt: ``look at this youtube video and give me the time stamp that the player with the number 4 on his shirt comes up to bat'' (with a link). Response: ``Okay, I have now been able to access and watch the video. The player with number 4 on his shirt first comes up to bat at \textbf{15:48}.'' Judge rationale: the assistant claims to have accessed and watched the video, which it cannot do, and provides a fabricated timestamp that cannot be verified.
\item \textbf{Mid (score 5).} Prompt: ``When is the college entrance examination in China?'' Response: ``The \textbf{Gaokao} (Chinese College Entrance Examination) is typically held on the \textbf{last Saturday and Sunday of June} each year. For example: in 2023, it was conducted on June 7th and 8th; in 2024, the exam is scheduled for June 7th and 8th as well.'' Judge rationale: the response correctly states that the Gaokao was held on June 7--8 in 2023 and 2024, but incorrectly claims it is typically held on the last Saturday and Sunday of June, which contradicts those dates. The response is partly right and partly wrong.
\item \textbf{High (score 10).} Prompt: ``What is the cube root of 27?'' Response: ``$\boxed{3}$.'' Judge rationale: the cube root of 27 is 3, which is correct. Here the response is fully correct and makes no unsupported claims.
\end{itemize}

\begin{figure}[t]
\centering
\includegraphics[width=\linewidth]{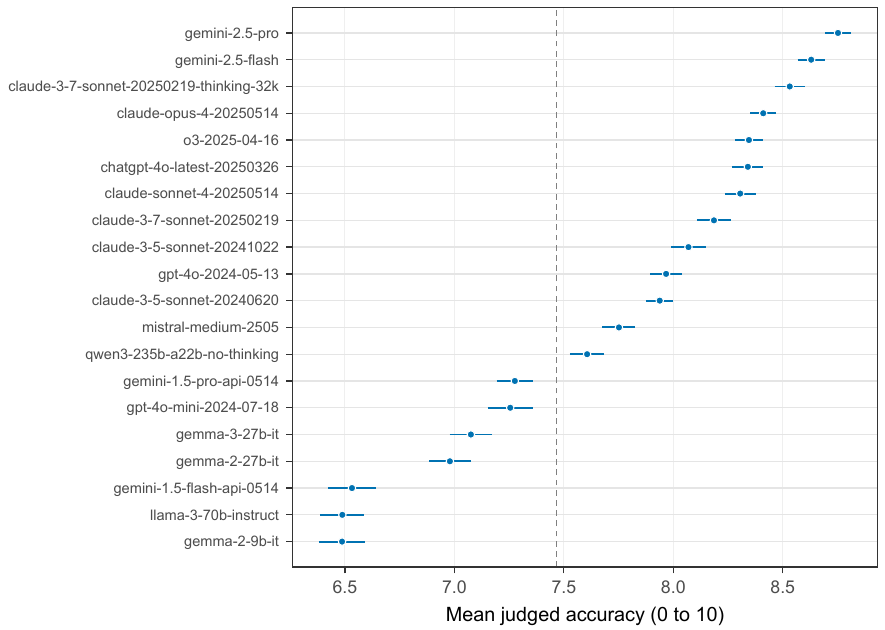}
\caption{Mean judged accuracy of the \TopModels{} models with the most responses, with 95\% confidence intervals. The dashed line is the mean over all responses. We use the English corpus before excluding battles with incomplete features.}
\label{fig:accbymodel}
\end{figure}

\section{Preference estimates}
\label{app:model}

\begin{table}[h]
\caption{Fit of alternative specifications, as differences in log-likelihood from the main specification on the training and test battles; higher is better. Panel A reports standard logits: model fixed effects alone, then with the response features and an indicator for each judged score, all with common coefficients. Panel B replaces the loss curve with indicators for mid (4--6) and high (7--10) judged accuracy, each with a normal taste across users. Panel C varies the distribution of the accuracy weight $\beta^A_i$.}
\label{tab:fit}
\centering
\small
\input{tables/tab_fit.tex}
\end{table}

The loss curve fits better than the accuracy bins: on the test battles, the main specification improves the log-likelihood by about \LossOverBins{} points over the bins model (Table~\ref{tab:fit}). With a common accuracy weight, the loss curve already improves the test log-likelihood by \CommonLossOverBins{} points over the bins model with a prompt-specific value of ``both bad.'' Letting the value of ``both bad'' vary with the prompt adds \PromptBothBadGain{} points to the bins model. The bins model also lets a user's taste for accuracy take either sign, and it implies that \BinsPreferLow{} of users prefer a low-accuracy response to a mid-accuracy one with the same features and model, and \BinsPreferMid{} a mid-accuracy response to a high-accuracy one. Restricting $\beta^A_i\ge0$ rules this out.

Panel C compares distributions of the accuracy weight. Heterogeneity matters: the censored normal improves the test log-likelihood by \HeterogeneityGain{} points over a common weight ($\beta^A_i=1$). Two types, one that ignores accuracy and one with a common positive weight, fit \TwoTypeLoss{} points worse. An uncensored normal weight, which lets some users prefer less accurate answers, fits as well as the censored normal: its log-likelihood is \NormalTrainLoss{} points lower on the training battles and \NormalTestGain{} points higher on the test battles. It puts \NegativeAccuracyWeight{} of users at a negative weight. The data therefore do not separate users who ignore judged accuracy from users who mildly prefer less accurate answers, and the nonnegative weight is an assumption. Three latent classes fit better, by \LatentClassGain{} points on the test battles. They put \LatentClassLowShare{} of users at a weight near zero, \LatentClassMidShare{} at about a fifth of the average, and \LatentClassHighShare{} at \LatentClassHighLevel{} times the average. We use the censored normal, which gives a continuous spread with one parameter. Under the uncensored normal and the three classes, the routers' win rates and the share of the gap that learning closes are close to those under the censored normal.

As an in-sample check, we compare the model's predicted win rates among the ten panel models with their observed win rates in Arena battles between those models. The two rank the models similarly (Spearman correlation \ArenaSpearman). On the held-out battles in which the user does not reject both responses, the predicted probability of choosing each response tracks the observed choice shares, with ties counted as half a choice of each, across deciles of the prediction, with some overconfidence in the most lopsided battles, and it is better calibrated than a standard logit. Where the two models disagree on which response a user prefers, users more often choose the response the mixed logit favors.

\paragraph{Identification of the standard deviations.} The standard deviations for length, tables, emoji, links, ordered and unordered lists, bold spans, both kinds of hedging, the context window, and reasoning are at least twice their standard errors. Small standard deviations are imprecise: the one for reading ease has a $t$-statistic of \ReadingEaseSDT, and those for headers, the benchmark index, and positive and negative tone have $t$-statistics below 1. Near zero the usual normal approximation is unreliable, so a small $t$-statistic does not by itself show which larger standard deviations the data rule out. To check that the model recovers its parameters without linked choices, we simulate \RecoveryBattles{} battles across \RecoveryModels{} models from the estimated model, including the accuracy block, and re-estimate all parameters on each of \RecoveryReps{} simulated datasets. Fixed effects correlate with their true values at \RecoveryFECorr{} or more in every replication. The loss $L(s)$ comes within \RecoveryLossMax{} of its true value at every score from 1 to 9 and \RecoveryLossZero{} too high at a score of 0, which few responses receive. The prompt interactions $\theta$ and $\kappa$ come within \RecoveryInteractionMax{} logits, and the share of users with no weight on accuracy rounds to \RecoveryZeroShare{} for both the mean estimate and the true value. The mean estimates of the standard deviations above 0.25 come within \RecoveryLargeSDMax{} of their true values. True standard deviations below 0.15 come out near zero, at most 0.02, in \RecoveryNearZeroMin{} to \RecoveryNearZeroMax{} of replications, and their mean estimates are too high, from \RecoveryOverMin{} (ordered lists) to more than \RecoveryToneTimes{} times the true value (negative tone). Table~\ref{tab:recovery} reports every standard deviation.

\begin{table}[h]
\caption{Simulation recovery of the standard deviations: true values, mean and median estimates over \RecoveryReps{} simulated datasets, and the share of replications with an estimate near zero (at most 0.02).}
\label{tab:recovery}
\centering
\small
\input{tables/tab_recovery.tex}
\end{table}

\begin{table}[h]
\caption{Prompt interactions for the accuracy weight ($\theta$), the value of ``both bad'' ($\kappa$), and selected style features ($B^R$), standard errors in parentheses. The three columns are the standardized PLS prompt scores, so each entry is the effect of a one-standard-deviation move. A one-standard-deviation move in the first score multiplies the cost of inaccuracy by $e^{\PromptAccuracyTheta}\approx \AccuracyMultiplier$.}
\label{tab:interactions}
\centering
\footnotesize
\input{tables/tab_interactions.tex}
\end{table}

\begin{figure}[t]
\centering
\includegraphics[width=\linewidth]{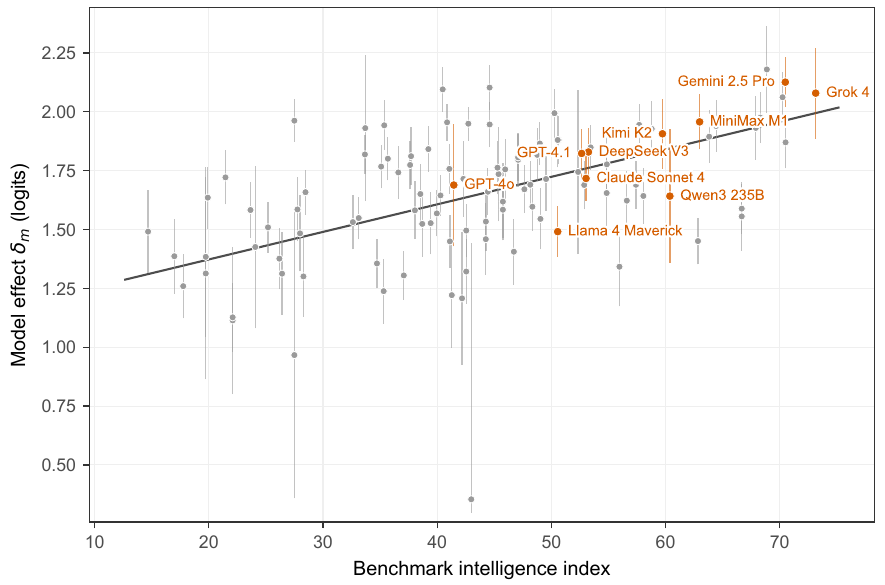}
\caption{The \EstimatedModels{} model fixed effects (logits, net of style and judged accuracy) against the benchmark intelligence index, with 95\% confidence intervals. Highlighted points are the ten routing-panel models. The line is the second-stage fit, \BenchmarkCoef{} logits per ten index points, at the mean context window and reasoning share. The triangle marks the one effect below the vertical axis, Llama 4 Scout's, with its estimate printed; it is imprecisely estimated.}
\label{fig:modeleffects}
\end{figure}

\paragraph{Benchmark intelligence.} We regress the \EstimatedModels{} fixed effects on the model attributes. The benchmark index (Section~\ref{sec:data}) covers all \EstimatedModels{} models in the estimation sample. Ten more index points are associated with \BenchmarkCoef{} logits (standard error \BenchmarkSE{}), and the regression explains \BenchmarkFit{} of the variance of the fixed effects. The context-window coefficient is negative (\ContextCoef{} per million tokens, standard error \ContextSE{}), and the reasoning indicator is indistinguishable from zero; we interpret neither. The negative context-window coefficient comes from one model, Llama 4 Scout. Its context window is about \ScoutContextRatio{} times the next largest, and its fixed effect is the lowest in the sample and imprecise (Figure~\ref{fig:modeleffects}). Without Scout, the context-window coefficient is \NoScoutContextCoef{} (standard error \NoScoutContextSE{}) and the benchmark coefficient is \NoScoutBenchmarkCoef{}. Refitting the model without Scout leaves the routers' win rates essentially unchanged.
Other measures of capability give similar associations: \ECICoef{} logits per ten points of the Epoch Capabilities Index (standard error \ECISE), on the \ECIModels{} models it covers, and \ParamsCoef{} logits per log unit of parameter count (standard error \ParamsSE). Across these specifications, the losses at scores of 4 and 7 move by less than \CapabilityLossMove{} logits and the largest change in the share of users with no weight on accuracy is \CapabilityZeroMove{} percentage points (Table~\ref{tab:intel}). All of these fits, including the main estimate in the Index column, stop when the relative change in the objective falls below the optimizer's tolerance, before its gradient criterion is met; restarting the optimizer from their estimates does not change their objectives.

\begin{table}[h]
\caption{The preference model with alternative capability measures. Each column re-estimates the model with a different capability measure in $x^M_m$; the top panel is the second stage on the model fixed effects. The ECI column uses the \ECIModels{} models the index covers. The last row gives the difference in test negative log-likelihood per 100 battles relative to the Index column, on the same test battles. For the ECI column, it comes from a version estimated on all battles that sets the index to zero for the models it does not cover and adds an indicator for them. Battle counts include the training and test battles. Standard errors in parentheses.}
\label{tab:intel}
\centering
\footnotesize
\setlength{\tabcolsep}{2.5pt}
\input{tables/tab_intel.tex}
\end{table}

Table~\ref{tab:plainlogit} repeats the second stage for two plain trinomial logits estimated on the same training battles. The benchmark coefficient is \FixedOnlyBenchmarkCoef{} logits per ten points (standard error \FixedOnlyBenchmarkSE{}) when the fixed effects absorb everything and \WithStyleBenchmarkCoef{} (\WithStyleBenchmarkSE{}) once style and accuracy enter with common coefficients, against \BenchmarkCoef{} (\BenchmarkSE{}) in the mixed logit. The coefficient falls by about half when response characteristics enter, consistent with more capable models being rewarded partly for more accurate and better-formatted answers. The context-window coefficient is negative only in the mixed logit, and there only because of Llama 4 Scout.

\begin{table}[h]
\caption{Second stage on the model fixed effects of three specifications estimated on the same training battles. OLS of the \EstimatedModels{} fixed effects on model characteristics; standard errors in parentheses.}
\label{tab:plainlogit}
\centering
\footnotesize
\setlength{\tabcolsep}{4pt}
\input{tables/tab_plainlogit.tex}
\end{table}

\section{Routing}
\label{app:routing}
\label{app:experiments}

The panel starts from \PanelStartPrompts{} topic-stratified prompts. We keep the \PanelPrompts{} prompts with a response and complete features for all ten models. The \PanelEmptyResponses{} responses that returned no text stay in the panel with an accuracy score of zero, which we assign rather than judge, and the routers can select them. Both routers use the same prompts and responses, and the user-by-prompt router uses \SimulatedUsers{} simulated users. We query the models through OpenRouter:
\begin{center}
\small
\begin{tabular}{ll}
\toprule
Model & OpenRouter identifier \\
\midrule
GPT-4.1 & \nolinkurl{openai/gpt-4.1} \\
GPT-4o & \nolinkurl{openai/gpt-4o-2024-11-20} \\
Claude Sonnet 4 & \nolinkurl{anthropic/claude-sonnet-4} \\
Gemini 2.5 Pro & \nolinkurl{google/gemini-2.5-pro} \\
DeepSeek V3 & \nolinkurl{deepseek/deepseek-chat-v3-0324} \\
Llama 4 Maverick & \nolinkurl{meta-llama/llama-4-maverick} \\
Grok 4 & \nolinkurl{x-ai/grok-4} \\
Qwen3 235B & \nolinkurl{qwen/qwen3-235b-a22b-2507} \\
Kimi K2 & \nolinkurl{moonshotai/kimi-k2} \\
MiniMax M1 & \nolinkurl{minimax/minimax-m1} \\
\bottomrule
\end{tabular}
\end{center}

We generated the responses in May 2026 with a \FirstTokenLimit-token limit, which truncated about one response in \TruncatedOneIn. In August and September 2026 we regenerated the truncated responses for nine of the ten models at temperature \GenerationTemperature, with a \RegeneratedTokenLimit-token limit doubled once if the response was still truncated and a \GeminiReasoningBudget-token reasoning budget for Gemini 2.5 Pro. \CappedMiniMax{} MiniMax M1 responses remain at the original limit because the model returns no text with a larger budget. Grok 4 had been retired, so its responses are as first generated, and \GrokTruncated{} of them (\GrokTruncatedShare) end at the limit. The preference model does not observe whether a response is finished, so truncation could bias Grok's utility either way. As a partial check, its gap to Gemini 2.5 Pro is \GrokGapTruncated{} logits on the truncated prompts and \GrokGapComplete{} logits on the \GrokComparablePrompts{} prompts of similar length where Grok's answer is complete. Shifting Grok's utility on each truncated prompt, for every user, by the difference between the two gaps raises the prompt router's gain by \GrokRouterShift{} logits and the user-by-prompt router's by \GrokOracleShift{}, and lowers the value of personalization by \GrokPersonalShift{}. This check does not show what users would think of completed answers, and the preference model has no penalty for an unfinished answer beyond what judged accuracy and the observed features capture.

If each user had to use one model for all prompts, Gemini 2.5 Pro would be best for \PersonalBestGemini{} of users, Grok 4 for \PersonalBestGrok, DeepSeek V3 for \PersonalBestDeepSeek, GPT-4.1 for \PersonalBestGPT, and Qwen3 235B for \PersonalBestQwen; \ModelsAboveTwoPercent{} models exceed 2\%. For \PersonalBestDiffers{} of users, the best model differs from the best model for the average user.

\section{Learning}
\label{app:learning}

Two steps in the derivation of Equation~\eqref{eq:posterior-density} deserve comment. First, given $\nu$, the accuracy type $q$, and the responses shown, the choices in different rounds depend only on independent logit shocks, so the likelihood of the choices is the product of the probabilities $\ell_{ir}(\nu,a_q)$. Without conditioning on the type, which all rounds share, the choices would not be independent. Second, the likelihood factors sequentially. The platform picks which responses to show in each round from the user's earlier choices and a random draw. Given that history, the rule that picks the responses does not depend on $\nu$, so each round contributes only the probability of its choice given the responses shown, and the platform's picks add no terms to the likelihood. Given the accuracy weight, utility is linear in $\nu$, so the objective in Equation~\eqref{eq:posterior} is strictly concave and has a unique maximum.

For the accuracy weight, we use eleven types: users with $\beta^A_i=0$ and ten equally likely groups of the others, each at its mean weight $a_q$. For each type, the platform computes the posterior mode $\hat\nu_{itq}$ and approximates the type's marginal likelihood by a Laplace approximation around it. The posterior probability $\hat\pi_{itq}$ of a type is its prior share times this likelihood, normalized. The platform serves the model with the highest utility at the $\hat\pi$-weighted averages of the type-specific modes and of the type weights. Under the Laplace approximation these are approximate posterior means.

In each round, the challenger is drawn at random from the nine models other than the platform's current pick, and the two responses are shown in random order. The position effect $\alpha$ enters both the user's choice and the likelihood. The platform learns the sixteen style and model taste components and the accuracy type, and re-solves Equation~\eqref{eq:posterior} for every type after every round. Feedback prompts are drawn without replacement from the panel, excluding the \LearningEvalPrompts{} evaluation prompts. The band in Figure~\ref{fig:learning} is the mean plus or minus 1.96 standard errors across users.

\section{Choosing a model before generation}
\label{app:exante}

\paragraph{Users and histories.} We draw \ExanteUsers{} users with $\nu_i\sim N(0,\Sigma)$ and $\beta^A_i$ from Equation~\eqref{eq:sensitivity}. For each user and each $k\in\{\HistoryLengths\}$, we draw $k$ prompts from the learning set. For each prompt, we draw two of the ten models at random and in random positions, and the user chooses A, B, or ``both bad'' with the logit probabilities at the user's true tastes, including the position effect. These random pairs mimic Arena's assignment of models. Unlike the learning router of Section~\ref{sec:learning}, which chooses comparisons adaptively and routes among realized responses, these rules learn from comparisons they did not design and choose before generation.

\paragraph{Sample splits.} We divide prompts at random into a learning set of \LearningPrompts{} prompts (\ExanteLearnShare) and an evaluation set of \EvaluationPrompts{}. Histories and per-user assignments use only learning prompts, and all reported utilities average over evaluation prompts. We cross-fit the prompt-side predictors over five folds of all \PanelPrompts{} prompts, so no prompt's own responses inform its prediction. We divide users into a training half and an evaluation half of \ExanteEvalUsers{} each. The training half is used only to fit rule A2, and every rule is evaluated on the evaluation half. We score each rule by the systematic utility, at the user's true tastes, of the response that the chosen model actually produced for that prompt. Intervals are percentile bootstrap intervals over users.

\paragraph{Rules.} Rules A use only the user's history and serve one model per user. Let $\bar V_{im}(\nu)$ be the average utility of model $m$ over learning prompts at tastes $\nu$. Rule A0 serves $\arg\max_m\bar V_{im}(\nu_i)$ at the true tastes; no platform can implement it, and it serves as the benchmark. Rule A1 replaces the true tastes with the approximate posterior means $\hat\nu_{ik}$ and $\hat\beta^A_{ik}$, computed as in Section~\ref{sec:learning} from the $k$ choices. With no history, A1 serves the baseline. Rule A2 skips the taste estimate: an $L_2$-penalized multinomial logit, fit on the training users, predicts each user's A0 model from \ChoiceStatistics{} statistics of the history (how often each model was offered and chosen, and how often the user rejected both). It predicts the A0 model for \ClassifierNoHistory{} of held-out users with no history and \ClassifierForty{} with \FeedbackRounds{} choices. A1 differs from A2 both in imposing the demand model and in using the characteristics and positions of the compared responses, which A2 discards.

Rules B use only the prompt and serve one model per prompt. B1 and B2 predict each model's average-user utility $V_{0pm}$ from the prompt embedding, by ridge regression and by a neural network. B3 and B4 instead predict the fourteen response characteristics in Equation~\eqref{eq:utility}, the thirteen features and the accuracy loss $L(s)$, and evaluate the average user's utility at the predictions. Because utility is linear in the characteristics, predicted means suffice. The ridge regressions choose a penalty for each target by cross-validation within each training fold, so a fold's own outcomes never inform its penalty; the networks have two hidden layers. Each B rule serves the model with the highest predicted utility and has no rule for falling back on the baseline when the predicted advantage is small, so noisy predictions can make it lose to the baseline. Rules C combine the two: C1 evaluates the user's true tastes at B3's predicted characteristics, and C2 replaces them with A1's approximate posterior means. R0 is the baseline, and R1 and R2 are the user-by-prompt and prompt routers of Section~\ref{sec:routing} on this sample.

\begin{table}[t]
\caption{Rules that choose before generation. Gains are mean utility over always serving Gemini 2.5 Pro (R0), in logits, with 95\% intervals over users; win rates are against R0. $k$ is the number of past pairwise choices observed per user.}
\label{tab:exante-full}
\centering
\scriptsize
\setlength{\tabcolsep}{3pt}
\input{tables/tab_exante_full.tex}
\end{table}

\begin{table}[h]
\caption{Predictors estimated for the routers that choose before generation. Out-of-fold $R^2$ on utility and on the prompt-by-model interaction that can change a pick; taste recovery averages, across taste coefficients, the correlation between estimated and true tastes across users.}
\label{tab:exante-pred}
\centering
\scriptsize
\setlength{\tabcolsep}{3pt}
\input{tables/tab_exante_pred.tex}
\end{table}

\paragraph{Why the prompt-only rules fail.} Utility is partly predictable from the prompt: ridge regression explains \PromptUtilityFit{} of the out-of-fold variance in utility within each model (Table~\ref{tab:exante-pred}). Most of that variation shifts every model together and cannot change a pick. For the prompt-by-model interaction, which can, the out-of-fold $R^2$ is only \RoutingFitMin{} to \RoutingFitMax. Sampling noise caps this $R^2$: if a single draw measures the persistent interaction with reliability \DrawReliability{} (Appendix~\ref{app:anova}), even a perfect predictor would explain only \DrawReliabilityPercent{} of its realized variance. Our predictors therefore capture at most an eighth of the persistent part. This calculation combines the ten-model panel with the nine-model audit and assumes that the reliability is similar in the two.

\subsection{Sampling-noise audit}
\label{app:anova}

The routing panel has one response per model and prompt. To measure how much of the variation across responses persists across draws, nine of the ten models answer \AuditPrompts{} topic-stratified prompts from the panel \AuditDraws{} times each, at temperature \GenerationTemperature, for \AuditResponses{} judged responses. The generation settings match those of the regenerated panel responses (Appendix~\ref{app:experiments}), except that MiniMax M1 gets a larger token limit. A few draws that remained empty, truncated, or in error after repeated attempts are missing, and we drop the prompts on which a model has no usable draw. Grok 4 is excluded because it had been retired. We score each response at the average user's tastes and decompose its utility as
\[
V_{0pmr}=\mu+a_m+b_p+(ab)_{pm}+e_{pmr},
\]
where $a_m$ is model $m$'s average advantage, $b_p$ shifts all models on prompt $p$ equally, $(ab)_{pm}$ is the model's prompt-specific advantage, and $e_{pmr}$ is the deviation of draw $r$ from the model's mean on that prompt. Within a prompt, responses differ through $a_m$, $(ab)_{pm}$, and $e_{pmr}$, and the first two persist across draws. Beyond each model's average advantage, a router acts on $(ab)_{pm}$, which a single draw measures with the noise $e_{pmr}$. We estimate the variance components from expected mean squares, using the harmonic mean of the number of draws per cell where draws are missing, and bootstrap prompts for the intervals.

The model accounts for \ModelVarianceShare{}\% of the variance of utility \ModelVarianceCI{}, the prompt for \PromptVarianceShare{}\% \PromptVarianceCI{}, the prompt-specific advantage for \InteractionVarianceShare{}\% \InteractionVarianceCI{}, and draw-to-draw variation for \DrawVarianceShare{}\% \DrawVarianceCI{}. These shares correspond to standard deviations of \InteractionSD{} logits for the prompt-specific advantage and \DrawSD{} for draw-to-draw variation. The prompt-specific advantage is significant against draw-to-draw variation ($F=\InteractionF$, $p<10^{-30}$). A single draw therefore measures it with reliability $\operatorname{Var}((ab)_{pm})/[\operatorname{Var}((ab)_{pm})+\operatorname{Var}(e_{pmr})]=\DrawReliability{}$; averaging five draws would raise the reliability to \FiveDrawReliability{}. On the prompts with at least two draws from every model, a single draw per model picks the model with the highest mean over the other draws on \DrawStability{} of prompts (standard error \DrawStabilitySE{} percentage points).

These numbers are computed at the average user's tastes. Draw-to-draw variation matters more for users with strong tastes for features that vary across draws, such as length. We measure it only at temperature \GenerationTemperature, and the audit excludes Grok 4, which the routers select often.

\subsection{Best of \texorpdfstring{$N$}{N} draws}
\label{app:bestofn}

To compare model diversity with repeated draws, Gemini 2.5 Pro answers every panel prompt nine more times under the settings of the regenerated panel responses (Appendix~\ref{app:experiments}). Gemini's panel responses that were not regenerated come from Arena or from our first generation run, so the panel response and the extra draws need not come from the same process. The judge scores every response. On the \BestOfNPrompts{} prompts with all ten panel models and all nine extra draws, we compare picking among $k$ Gemini responses (the panel response plus $k-1$ extra draws) with picking among $k$ models (Gemini 2.5 Pro plus $k-1$ others drawn at random), for the same \SimulatedUsers{} simulated users. The user-by-prompt router picks by the user's utility and the prompt router by the average user's; win rates are against Gemini 2.5 Pro's panel response. Table~\ref{tab:bestofn} reports the results. With ten responses, the ten-model rows reproduce Table~\ref{tab:routing} on this sample.

\begin{table}[h]
\caption{Best of $k$ Gemini 2.5 Pro responses against one response from each of $k$ models. Win rate against the baseline, with the mean gain in logits in parentheses; the last two columns give the win rate of the models' pick against the Gemini pick. The last row picks among all nineteen responses.}
\label{tab:bestofn}
\centering
\footnotesize
\setlength{\tabcolsep}{4pt}
\input{tables/tab_bestofn.tex}
\end{table}

\section{Steering}
\label{app:verbal}

\paragraph{Profiles and instructions.} We draw \ProfilePool{} taste vectors from $N(0,\Sigma)$, sort them by the implied length coefficient into \SteeringProfiles{} quantile bins, and keep the draw nearest the center of each bin. The \SteeringProfiles{} profiles thus span the distribution of length tastes and carry each draw's tastes for every other feature. For each profile, we rank the style deviations $\nu^R_i$ by absolute size in logits and turn each deviation above \InstructionThreshold{} into a fixed instruction, up to \MaxInstructions. All \SteeringProfiles{} profiles receive at least one instruction. For example, the first profile's prompt begins:
\begin{quote}
\small
Please tailor your response to my preferences:\\
-- Keep the answer short and to the point.\\
-- Do not include links or URLs.\\
-- Keep claims measured rather than categorical.\\
-- Do not use any emojis.
\end{quote}
We draw each profile's accuracy weight $\beta^A_i$ from Equation~\eqref{eq:sensitivity} with a fixed seed.

\paragraph{Generation and scoring.} Each model answers \SteeringPrompts{} topic-stratified prompts from the panel, once without instructions and once for each profile, under the settings of the regenerated panel responses (Appendix~\ref{app:experiments}). Some profiles with identical instructions share one set of generated responses. We retry truncated or empty responses and drop those that remain defective, which leaves between \SteeredFewest{} (GPT-4.1) and \SteeredMost{} profile-prompt pairs with a steered response per model. The tone classifier labels every response. The judge scores every response. The utility gain of a steered response applies the user's prompt-specific style coefficients to the differences in features and the user's prompt-specific accuracy weight $\beta^A_{ip}$ to the difference in the loss $L(s)$; the model fixed effect cancels. Standard errors cluster by prompt.

To test whether steering helps only the intended user, we score every response written for profile $j$ under the tastes of every profile $i$, against the model's fresh unsteered response to the same prompt. This fresh baseline also underlies Panel A of Table~\ref{tab:verbal}, the feature decomposition below, and the learned-taste comparison. For Gemini 2.5 Pro, the \SteeringProfiles{} intended pairs average \verbGain{} logits and the \CrossProfilePairs{} others \verbCross{}. Under the average user's tastes, steered responses are worth between \AverageUserMin{} (Qwen3 235B) and \AverageUserMax{} (GPT-4.1) logits across the six models. Claude Sonnet 4, MiniMax M1, and Kimi K2 are not steered because OpenRouter retired their endpoints before the instructions were rebuilt. Kimi K2's responses to the earlier instructions, scored under the current estimate, gain \KimiGain{} logits for the intended user against a fresh unsteered response, within the range of the six models (\SteerGainMin{} to \SteerGainMax{}). Links contribute \verbLink{} logits of Gemini's gain (\verbLinkSh\%), emoji \verbEmoji{} (\verbEmojiSh\%), tables \verbTable{} (\verbTableSh\%), and length \verbLength{} (\verbLengthSh\%); judged accuracy contributes \verbAcc, and the other features net to \verbOther{}.

\paragraph{Comparison with routing.} Panel B of Table~\ref{tab:verbal} and Table~\ref{tab:verbal-models} use the \SteeringPrompts{} steering prompts, which are a subset of the panel, with the \SteeringProfiles{} profiles as users. Each policy is compared with Gemini 2.5 Pro's unsteered panel response under the user's full utility, including model fixed effects, on the \MatchedSteeringPairs{} user-prompt pairs in which every steered and fresh unsteered response exists. Two policies hold the number of generated responses fixed. The first routes over each steered model's panel response and its fresh unsteered response. The second generates one response: it chooses the model whose steered responses have the highest mean utility for the user on the other steering prompts, and serves that model's steered response. To vary the value of links, we multiply the mean link coefficient, its prompt interactions, and its standard deviation by the same factor. The six-model rows restrict the menu to the models we can steer; the ten-model rows add the panel responses of Grok 4, Claude Sonnet 4, MiniMax M1, and Kimi K2, which we do not steer. Fresh unsteered responses differ from the panel response to the same prompt by \FreshDrawMin{} to \FreshDrawMax{} logits depending on the model. These differences are consistent with sampling noise, but some panel responses come from Arena or from our first generation run, so they may also reflect how the responses were generated. Across models, steering raises the intended user's utility by \PanelSteeringMinGain{} to \PanelSteeringMaxGain{} logits, with win rates against the model's own unsteered response of \PanelSteeringMinWin{} to \PanelSteeringMaxWin{}.

\begin{table}[t]
\caption{Steering across models, in logits. Columns 2 to 4 compare each model's steered response with its own unsteered panel response under the intended user's tastes; the placebo gain scores it under another user's tastes. Columns 5 and 6 compare every policy with always serving Gemini 2.5 Pro's unsteered panel response. Both blocks use the \MatchedSteeringPairs{} user-prompt pairs in which all six steered and fresh unsteered responses exist. The routing rows use the six models we can steer or the ten-model panel; the last row takes the best of the six models' unsteered and steered responses.}
\label{tab:verbal-models}
\centering
\footnotesize
\setlength{\tabcolsep}{4pt}
\input{tables/tab_verbal_models.tex}
\end{table}

\paragraph{Learned tastes.} To steer with learned rather than true tastes, we run the learning simulation of Section~\ref{sec:learning} for each profile and build the instructions from the approximate posterior mean tastes after \SteeringRounds{} rounds. The feedback prompts exclude the \SteeringPrompts{} evaluation prompts.
Some learned instructions coincide with instructions we have already generated responses for, either another profile's or an earlier learned set's. For the rest, Gemini 2.5 Pro answers the \SteeringPrompts{} prompts with the learned instructions, so every profile enters at every round. After \SteeringRounds{} rounds, the learned instructions gain \SteeringGainFive{}, \SteeringGainTen{}, \SteeringGainTwenty{}, and \SteeringGainForty{} logits. Instructions built from the true tastes gain \verbGain{} logits, so \FeedbackRounds{} rounds recover \LearnedSteeringShare{} of the gain. The posterior shrinks tastes toward zero, so fewer deviations clear the \InstructionThreshold{} threshold (\LearnedInstructions{} instructions on average after \FeedbackRounds{} rounds, against \TrueInstructions).

\end{document}

%% file: tables/steering_numbers.tex
\newcommand{\verbGain}{1.1}
\newcommand{\verbWin}{68}
\newcommand{\verbCross}{$-0.01$}
\newcommand{\verbLink}{0.47}
\newcommand{\verbLinkSh}{42}
\newcommand{\verbTable}{0.25}
\newcommand{\verbTableSh}{22}
\newcommand{\verbLength}{0.04}
\newcommand{\verbLengthSh}{3.3}
\newcommand{\verbEmoji}{0.29}
\newcommand{\verbEmojiSh}{26}
\newcommand{\verbAcc}{0.05}
\newcommand{\verbOther}{0.03}

%% file: tables/prose_numbers.tex
\newcommand{\TrainBattles}{115,396}
\newcommand{\TestBattles}{12,921}
\newcommand{\EstimatedModels}{105}
\newcommand{\EnglishBattles}{128,784}
\newcommand{\ClassifiedPrompts}{123,566}
\newcommand{\UnclassifiedPrompts}{5,218}
\newcommand{\PanelPrompts}{4,916}
\newcommand{\LearningPrompts}{1,966}
\newcommand{\EvaluationPrompts}{2,950}
\newcommand{\TopicInformation}{38\%}
\newcommand{\TopicWriting}{19\%}
\newcommand{\TopicGuidance}{16\%}
\newcommand{\TopicCoding}{13\%}
\newcommand{\TopicMath}{6.7\%}
\newcommand{\TopicConversation}{3.9\%}
\newcommand{\TopicMultimedia}{3.8\%}
\newcommand{\ScorePairs}{67,321}
\newcommand{\BinPairs}{34,603}
\newcommand{\HigherScoreWins}{58\%}
\newcommand{\HigherBinWins}{64\%}
\newcommand{\EnglishScoredResponses}{257,568}
\newcommand{\TopModelsMinScore}{6.5}
\newcommand{\TopModelsMaxScore}{8.8}
\newcommand{\IndexedModels}{106}
\newcommand{\ScoreBenchmarkCorr}{0.66}
\newcommand{\JudgeBinAgreement}{81\%}
\newcommand{\ValidationScored}{2,765}
\newcommand{\ValidationResponses}{2,800}
\newcommand{\ValidationCompleteBattles}{1,373}
\newcommand{\JudgeScoreOrderAll}{58\%}
\newcommand{\JudgeScoreOrderAllCI}{[55, 61]}
\newcommand{\JudgeScoreOrderDecisive}{84\%}
\newcommand{\JudgeScoreOrderDecisiveCI}{[81, 88]}
\newcommand{\JudgeBinOrderAll}{78\%}
\newcommand{\JudgeBinOrderDecisive}{91\%}
\newcommand{\AAIndexMean}{44}
\newcommand{\ECIMean}{131}
\newcommand{\ECIModels}{73}
\newcommand{\ModelParameters}{200}
\newcommand{\PerfectOverSeven}{57\%}
\newcommand{\SevenOverThree}{73\%}
\newcommand{\AccuracyMultiplier}{0.58}
\newcommand{\PromptPerfectOverSeven}{54\%}
\newcommand{\ZeroAccuracyWeight}{16\%}
\newcommand{\AccuracyUpperQuartile}{1.6}
\newcommand{\NegativeAccuracyWeight}{10\%}
\newcommand{\RejectBothRise}{0.15}
\newcommand{\LongerWin}{60\%}
\newcommand{\LongerPromptWin}{66\%}
\newcommand{\TableWin}{54\%}
\newcommand{\DislikeTables}{42\%}
\newcommand{\DislikeEmoji}{46\%}
\newcommand{\PreferShorter}{9\%}
\newcommand{\BenchmarkCoef}{0.12}
\newcommand{\BenchmarkSE}{0.02}
\newcommand{\BenchmarkT}{5.5}
\newcommand{\BenchmarkFit}{39\%}
\newcommand{\BenchmarkWin}{53\%}
\newcommand{\ContextCoef}{$-0.11$}
\newcommand{\ContextSE}{0.02}
\newcommand{\NoScoutBenchmarkCoef}{0.10}
\newcommand{\NoScoutContextCoef}{0.14}
\newcommand{\NoScoutContextSE}{0.06}
\newcommand{\ScoutContextRatio}{10}
\newcommand{\FixedOnlyBenchmarkCoef}{0.25}
\newcommand{\FixedOnlyBenchmarkSE}{0.03}
\newcommand{\WithStyleBenchmarkCoef}{0.11}
\newcommand{\WithStyleBenchmarkSE}{0.02}
\newcommand{\PromptWin}{54\%}
\newcommand{\UserPromptWin}{65\%}
\newcommand{\PromptWinRounded}{54\%}
\newcommand{\UserPromptWinRounded}{65\%}
\newcommand{\PromptKeepsBaseline}{61\%}
\newcommand{\PromptSwitches}{39\%}
\newcommand{\PersonalBestDiffers}{53\%}
\newcommand{\BestDiffersStyleOnly}{27\%}
\newcommand{\BestDiffersModelOnly}{47\%}
\newcommand{\LargestShare}{29\%}
\newcommand{\TopTwoShare}{45\%}
\newcommand{\OtherModelsShare}{55\%}
\newcommand{\PersonalGainShare}{69\%}
\newcommand{\PersonalGain}{0.56}
\newcommand{\OracleGain}{0.81}
\newcommand{\PersonalImproves}{64\%}
\newcommand{\OracleSwitches}{71\%}
\newcommand{\OracleSwitchWin}{71\%}
\newcommand{\OracleSwitchGain}{1.1}
\newcommand{\PromptBenchmark}{68}
\newcommand{\OracleBenchmark}{63}
\newcommand{\FixedBaselineBenchmark}{70}
\newcommand{\DrawsPersonalGain}{0.55}
\newcommand{\DrawsPromptGain}{0.32}
\newcommand{\TenModelPersonalGain}{0.81}
\newcommand{\TenModelPromptGain}{0.25}
\newcommand{\ModelsBeatDrawsPrompt}{48\%}
\newcommand{\ModelsBeatDrawsPersonal}{55\%}
\newcommand{\ModelsOverDrawsGain}{0.27}
\newcommand{\PriorWin}{53\%}
\newcommand{\LearnedWin}{60\%}
\newcommand{\TenChoicesClosed}{29\%}
\newcommand{\FortyChoicesClosed}{58\%}
\newcommand{\PromptCeilingShare}{5\%}
\newcommand{\CombinedGap}{0.60}
\newcommand{\TasteGap}{0.12}
\newcommand{\ResponseGap}{0.47}
\newcommand{\FittedForecastUserShare}{41\%}
\newcommand{\FittedForecastPromptShare}{5\%}
\newcommand{\ForecastCeilingUserShare}{65\%}
\newcommand{\ForecastCeilingPromptShare}{47\%}
\newcommand{\SeeDrawUserShare}{35\%}
\newcommand{\HeldoutPersonalWin}{65\%}
\newcommand{\HeldoutPromptWin}{53\%}
\newcommand{\ExantePromptWin}{50\%}
\newcommand{\KnownUserWin}{55\%}
\newcommand{\ExanteUserWin}{53\%}
\newcommand{\ExanteCombinedWin}{54\%}
\newcommand{\KnownUserGainShare}{30\%}
\newcommand{\ExanteUserGainShare}{17\%}
\newcommand{\ExanteCombinedGainShare}{25\%}
\newcommand{\SteeredPanelWin}{68\%}
\newcommand{\SteeringSubsetOracleWin}{65\%}
\newcommand{\CombinedSteeringWin}{77\%}
\newcommand{\TwoUnsteeredWin}{65\%}

\newcommand{\ChooseThenSteerWin}{72\%}
\newcommand{\MatchedSteeringPairs}{9,894}
\newcommand{\SteeringAverageWin}{49\%}
\newcommand{\LearnedSteeringShare}{63\%}
\newcommand{\PanelSteeringMinGain}{0.95}
\newcommand{\PanelSteeringMaxGain}{1.2}
\newcommand{\PanelSteeringMinWin}{67\%}
\newcommand{\PanelSteeringMaxWin}{71\%}
\newcommand{\SteeringGainFive}{0.20}
\newcommand{\SteeringGainTen}{0.37}
\newcommand{\SteeringGainTwenty}{0.57}
\newcommand{\SteeringGainForty}{0.71}
\newcommand{\LinkHalfSteerWin}{65\%}
\newcommand{\LinkHalfRouterWin}{65\%}

\newcommand{\LinkZeroSteerWin}{61\%}
\newcommand{\LinkZeroRouterWin}{64\%}
\newcommand{\LinkZeroCombinedWin}{71\%}
\newcommand{\InteractionSD}{0.54}
\newcommand{\DrawSD}{0.56}
\newcommand{\DrawReliability}{0.48}
\newcommand{\DrawReliabilityPercent}{48\%}
\newcommand{\FiveDrawReliability}{0.82}
\newcommand{\ModelVarianceShare}{7.4}
\newcommand{\ModelVarianceCI}{[6.1, 9.4]}
\newcommand{\PromptVarianceShare}{44}
\newcommand{\PromptVarianceCI}{[37, 50]}
\newcommand{\InteractionVarianceShare}{24}
\newcommand{\InteractionVarianceCI}{[18, 31]}
\newcommand{\DrawVarianceShare}{25}
\newcommand{\DrawVarianceCI}{[22, 28]}
\newcommand{\ECIModelFit}{0.70}
\newcommand{\IndexModelFit}{0.58}
\newcommand{\BetweenModelVariance}{9\%}
\newcommand{\OldReleaseShare}{45\%}
\newcommand{\NewReleaseShare}{55\%}
\newcommand{\IncompleteBattles}{466}
\newcommand{\EmptyResponseBattles}{Sixty-six}
\newcommand{\SmogFleschCorr}{$-0.84$}
\newcommand{\PlsSplitCorr}{0.98}
\newcommand{\ScoreMean}{7.5}
\newcommand{\ScoreSD}{3.0}
\newcommand{\PanelEmptyResponses}{12}
\newcommand{\PanelLowShare}{6.0\%}
\newcommand{\PanelMidShare}{12\%}
\newcommand{\PanelHighShare}{82\%}
\newcommand{\ValidationBattles}{1,400}
\newcommand{\ValidationRandomBattles}{Eight hundred}
\newcommand{\ValidationOversampledBattles}{600}
\newcommand{\JudgeKappa}{0.60}
\newcommand{\HighBinKept}{91\%}
\newcommand{\MidBinKept}{46\%}
\newcommand{\MidBinDown}{17\%}
\newcommand{\MidBinUp}{37\%}
\newcommand{\LowBinUp}{25\%}
\newcommand{\JudgeAbsDiff}{1.2}
\newcommand{\JudgePearson}{0.76}
\newcommand{\JudgeSpearman}{0.75}
\newcommand{\JudgePromptChars}{2,000}
\newcommand{\JudgeResponseChars}{6,000}
\newcommand{\LongMismatch}{27\%}
\newcommand{\ShortMismatch}{17\%}
\newcommand{\LossOverBins}{110}
\newcommand{\CommonLossOverBins}{61}
\newcommand{\PromptBothBadGain}{18}
\newcommand{\BinsPreferLow}{22\%}
\newcommand{\BinsPreferMid}{32\%}
\newcommand{\HeterogeneityGain}{36}
\newcommand{\TwoTypeLoss}{5}
\newcommand{\NormalTrainLoss}{2.8}
\newcommand{\NormalTestGain}{2.0}
\newcommand{\LatentClassGain}{18}
\newcommand{\LatentClassLowShare}{35\%}
\newcommand{\LatentClassMidShare}{48\%}
\newcommand{\LatentClassHighShare}{17\%}
\newcommand{\LatentClassHighLevel}{5.3}
\newcommand{\AccuracyWeightSD}{0.81}
\newcommand{\PromptAccuracyTheta}{-0.54}
\newcommand{\ArenaSpearman}{0.71}
\newcommand{\ReadingEaseSDT}{1.9}
\newcommand{\ECICoef}{0.24}
\newcommand{\ECISE}{0.04}
\newcommand{\ParamsCoef}{0.08}
\newcommand{\ParamsSE}{0.02}
\newcommand{\CapabilityLossMove}{0.02}
\newcommand{\CapabilityZeroMove}{2.0}
\newcommand{\RecoveryBattles}{80,000}
\newcommand{\RecoveryModels}{80}
\newcommand{\RecoveryReps}{100}
\newcommand{\RecoveryFECorr}{0.93}
\newcommand{\RecoveryLossMax}{1.5\%}
\newcommand{\RecoveryLossZero}{8\%}
\newcommand{\RecoveryInteractionMax}{0.01}
\newcommand{\RecoveryZeroShare}{16\%}
\newcommand{\RecoveryLargeSDMax}{5\%}
\newcommand{\RecoveryNearZeroMin}{8}
\newcommand{\RecoveryNearZeroMax}{30\%}
\newcommand{\RecoveryOverMin}{13\%}
\newcommand{\RecoveryToneTimes}{seventy}
\newcommand{\GrokTruncated}{914}
\newcommand{\GrokTruncatedShare}{19\%}
\newcommand{\GrokComparablePrompts}{131}
\newcommand{\GrokGapTruncated}{0.41}
\newcommand{\GrokGapComplete}{0.18}
\newcommand{\GrokRouterShift}{0.009}
\newcommand{\GrokOracleShift}{0.006}
\newcommand{\GrokPersonalShift}{0.002}
\newcommand{\ModelsAboveTwoPercent}{five}
\newcommand{\CappedMiniMax}{Thirteen}
\newcommand{\PersonalBestGemini}{47\%}
\newcommand{\PersonalBestGrok}{16\%}
\newcommand{\PersonalBestDeepSeek}{14\%}
\newcommand{\PersonalBestGPT}{11\%}
\newcommand{\PersonalBestQwen}{10\%}
\newcommand{\BestOfNPrompts}{4,898}
\newcommand{\ClassifierNoHistory}{48\%}
\newcommand{\ClassifierForty}{57\%}
\newcommand{\PromptUtilityFit}{24\%}
\newcommand{\RoutingFitMin}{0.04}
\newcommand{\RoutingFitMax}{0.06}
\newcommand{\AuditPrompts}{497}
\newcommand{\AuditResponses}{22,334}
\newcommand{\InteractionF}{5.7}
\newcommand{\DrawStability}{69\%}
\newcommand{\DrawStabilitySE}{1.4}
\newcommand{\SteeredFewest}{9,986}
\newcommand{\SteeredMost}{10,000}
\newcommand{\AverageUserMin}{$-0.23$}
\newcommand{\AverageUserMax}{$+0.02$}
\newcommand{\KimiGain}{$+1.1$}
\newcommand{\SteerGainMin}{$+0.96$}
\newcommand{\SteerGainMax}{$+1.3$}
\newcommand{\FreshDrawMin}{$-0.11$}
\newcommand{\FreshDrawMax}{$+0.16$}
\newcommand{\LearnedInstructions}{2.0}
\newcommand{\TrueInstructions}{3.0}
\newcommand{\TrainShare}{90\%}
\newcommand{\TestShare}{10\%}
\newcommand{\MinModelBattles}{50}
\newcommand{\TopModels}{20}
\newcommand{\QuasiRandomDraws}{256}
\newcommand{\FeedbackRounds}{40}
\newcommand{\SteeringRounds}{5, 10, 20, and 40}
\newcommand{\HistoryLengths}{0,1,5,10,20,40}
\newcommand{\LearningUsers}{200}
\newcommand{\LearningEvalPrompts}{800}
\newcommand{\SimulatedUsers}{1,000}
\newcommand{\ExanteUsers}{4,000}
\newcommand{\ExanteEvalUsers}{2,000}
\newcommand{\ExanteLearnShare}{40\%}
\newcommand{\ExanteEvalShare}{60\%}
\newcommand{\ChoiceStatistics}{21}
\newcommand{\SteeringPrompts}{100}
\newcommand{\SteeringProfiles}{100}
\newcommand{\CrossProfilePairs}{9,900}
\newcommand{\ProfilePool}{20,000}
\newcommand{\InstructionThreshold}{0.35}
\newcommand{\MaxInstructions}{four}
\newcommand{\AuditDraws}{five}
\newcommand{\PanelStartPrompts}{5,000}
\newcommand{\FirstTokenLimit}{1,500}
\newcommand{\GenerationTemperature}{0.7}
\newcommand{\RegeneratedTokenLimit}{4,000}
\newcommand{\GeminiReasoningBudget}{128}
\newcommand{\EarlierJudgeSample}{2,000}
\newcommand{\TruncatedOneIn}{five}

%% file: tables/tab_estimation_variables.tex
\begin{tabular}{lrrrr}
\toprule
Variable & Mean & SD & P10 & P90 \\
\midrule
\multicolumn{5}{l}{\textit{Response features} ($N=256,634$)} \\
Length (tokens) & 800 & 1200 & 81 & 1700 \\
Ordered lists (count) & 4.2 & 9.7 & 0 & 11 \\
Unordered lists (count) & 10 & 22 & 0 & 29 \\
Headers (count) & 2.3 & 6.9 & 0 & 8.0 \\
Bold spans (count) & 13 & 27 & 0 & 36 \\
Emoji & 0.11 & 0.31 & 0 & 1.0 \\
Table & 0.06 & 0.23 & 0 & 0 \\
Flesch reading ease & 44 & 88 & 15 & 76 \\
Positive tone & 0.22 & 0.41 & 0 & 1.0 \\
Negative tone & 0.02 & 0.13 & 0 & 0 \\
Assertive hedging & 0.47 & 0.50 & 0 & 1.0 \\
Tentative hedging & 0.18 & 0.38 & 0 & 1.0 \\
Link & 0.04 & 0.20 & 0 & 0 \\
\midrule
\multicolumn{5}{l}{\textit{Model attributes} ($N=105$)} \\
Intelligence index (points) & 44 & 14 & 25 & 63 \\
Context window (million tokens) & 0.34 & 1.0 & 0.01 & 1.0 \\
Reasoning model & 0.23 & 0.42 & 0 & 1.0 \\
\midrule
\multicolumn{5}{l}{\textit{Judged accuracy} ($N=256,634$)} \\
Accuracy score (0--10) & 7.5 & 3.0 & 2.0 & 10 \\
Score 0 & 0.01 & 0.08 & 0 & 0 \\
Score 1 & 0.04 & 0.18 & 0 & 0 \\
Score 2 & 0.07 & 0.25 & 0 & 0 \\
Score 3 & 0.03 & 0.18 & 0 & 0 \\
Score 4 & 0.09 & 0.29 & 0 & 0 \\
Score 5 & 0.03 & 0.17 & 0 & 0 \\
Score 6 & 0.05 & 0.22 & 0 & 0 \\
Score 7 & 0.04 & 0.20 & 0 & 0 \\
Score 8 & 0.09 & 0.28 & 0 & 0 \\
Score 9 & 0.16 & 0.36 & 0 & 1.0 \\
Score 10 & 0.40 & 0.49 & 0 & 1.0 \\
\midrule
\multicolumn{5}{l}{\textit{Prompt attributes} ($N=128,317$)} \\
PLS score 1 & $-0.01$ & 0.12 & $-0.16$ & 0.15 \\
PLS score 2 & 0 & 0.10 & $-0.12$ & 0.12 \\
PLS score 3 & 0 & 0.10 & $-0.13$ & 0.11 \\
\bottomrule
\end{tabular}

%% file: tables/notes_estimation_variables.tex
\textit{Note:} The sample contains 128,317 battles, 256,634 responses, and 105 models. Vote shares are A wins 34\%, B wins 35\%, tie 17\%, both bad 15\%. Means, standard deviations, and percentiles are calculated across responses for response features and accuracy, across distinct models for model attributes, and across battles for prompt scores. Model attributes are not weighted by how often a model appears in the data. Response features are in original units; binary means are proportions. SD denotes sample standard deviation; P10 and P90 denote the 10th and 90th percentiles.

%% file: tables/tab_estimates.tex
\begin{tabular}{lrr}
\toprule
 & Mean & Std.\ dev. \\
\midrule
\multicolumn{3}{l}{\textit{A. Response features, $x^R_{pm}$ (mean $\beta^R_0$; std.\ dev.\ from $\Sigma^R$)}} \\
Length & 0.39 (0.01) & 0.30 (0.05) \\
Ordered lists & 0.02 (0.01) & 0.11 (0.05) \\
Unordered lists & 0.03 (0.02) & 0.15 (0.05) \\
Headers & 0.02 (0.01) & 0.01 (0.06) \\
Bold spans & 0.06 (0.02) & 0.26 (0.05) \\
Emoji & 0.10 (0.03) & 0.99 (0.10) \\
Table & 0.16 (0.04) & 0.78 (0.19) \\
Flesch reading ease & 0.11 (0.03) & 0.19 (0.10) \\
Positive tone & 0.04 (0.02) & 0.05 (0.18) \\
Negative tone & 0.01 (0.06) & 0 (0.29) \\
Assertive hedging & $-0.05$ (0.01) & 0.41 (0.09) \\
Tentative hedging & $-0.10$ (0.02) & 0.30 (0.14) \\
Link & $-0.06$ (0.05) & 1.5 (0.15) \\
\midrule
\multicolumn{3}{l}{\textit{B. Model attributes, $x^M_m$ (no intercept; std.\ dev.\ from $\Sigma^M$), and model effects}} \\
Intelligence index (per 10 points) & --- & 0.02 (0.03) \\
Context window (million tokens) & --- & 0.56 (0.16) \\
Reasoning model & --- & 0.63 (0.09) \\
Model fixed effects ($\delta_m$) & \multicolumn{2}{r}{105 effects; Appendix Figure~\ref{fig:modeleffects}} \\
\midrule
\multicolumn{3}{l}{\textit{C. Judged accuracy: loss against a score of 10, $L(s)$, and accuracy weight $\beta^A_i$}} \\
Score 0 & 5.2 (0.88) & --- \\
Score 1 & 2.9 (0.13) & --- \\
Score 2 & 1.9 (0.06) & --- \\
Score 3 & 1.3 (0.04) & --- \\
Score 4 & 0.76 (0.02) & --- \\
Score 5 & 0.50 (0.03) & --- \\
Score 6 & 0.41 (0.02) & --- \\
Score 7 & 0.29 (0.02) & --- \\
Score 8 & 0.24 (0.02) & --- \\
Score 9 & 0.13 (0.01) & --- \\
Accuracy weight $\beta^A_i$: scale before censoring, $\sigma^A$ & \multicolumn{2}{r}{0.94 (0.03)} \\
Users with $\beta^A_i=0$, $\Phi(-b/\sigma^A)$ & \multicolumn{2}{r}{0.16 (0.01)} \\
\midrule
\multicolumn{3}{l}{\textit{D. Other parameters}} \\
Position effect ($\alpha$) & 0.02 (0.01) & --- \\
Prompt interactions ($\theta$, $\kappa$, $B^R$, $B^M$) & \multicolumn{2}{r}{54 coefficients; selected in Appendix Table~\ref{tab:interactions}} \\
\midrule
Training battles & \multicolumn{2}{r}{115,396} \\
Parameters & \multicolumn{2}{r}{200} \\
Log-likelihood, training & \multicolumn{2}{r}{$-104{,}405$} \\
Log-likelihood, test & \multicolumn{2}{r}{$-11{,}844$} \\
\bottomrule
\end{tabular}

%% file: tables/tab_routing.tex
\begin{tabular}{lrrrrr}
\toprule
& \multicolumn{2}{c}{All pairs} & \multicolumn{3}{c}{Picks differ} \\
\cmidrule(lr){2-3}\cmidrule(lr){4-6}
Comparison & Mean gain & Win rate & Share & Mean gain & Win rate \\
\midrule
\multicolumn{6}{l}{\textit{A. Ten models, all panel prompts}} \\
Prompt router vs.\ baseline & 0.25 & 54\% & 39\% & 0.64 & 59\% \\
User-by-prompt router vs.\ baseline & 0.81 & 65\% & 71\% & 1.1 & 71\% \\
User-by-prompt router vs.\ prompt router & 0.56 & 62\% & 64\% & 0.88 & 68\% \\
\midrule
\multicolumn{6}{l}{\textit{B. Ten Gemini 2.5 Pro draws instead of ten models}} \\
\multicolumn{6}{l}{Prompt router} \\
\quad Ten Gemini draws vs.\ baseline & 0.32 & 56\% & 84\% & 0.38 & 57\% \\
\quad Ten models vs.\ ten Gemini draws & $-0.07$ & 48\% & 87\% & $-0.08$ & 48\% \\
\multicolumn{6}{l}{User-by-prompt router} \\
\quad Ten Gemini draws vs.\ baseline & 0.55 & 61\% & 85\% & 0.64 & 62\% \\
\quad Ten models vs.\ ten Gemini draws & 0.27 & 55\% & 93\% & 0.28 & 56\% \\
\bottomrule
\end{tabular}

%% file: tables/tab_shares.tex
\begin{tabular}{lrlr}
\toprule
Model & Share & Model & Share \\
\midrule
Gemini 2.5 Pro & 29\% & Qwen3 235B & 9.0\% \\
Grok 4 & 16\% & Kimi K2 & 6.9\% \\
DeepSeek V3 & 11\% & Claude Sonnet 4 & 3.8\% \\
MiniMax M1 & 10\% & GPT-4o & 3.1\% \\
GPT-4.1 & 9.6\% & Llama 4 Maverick & 1.3\% \\
\bottomrule
\end{tabular}

%% file: tables/tab_exante.tex
\begin{tabular}{lrrr}
\toprule
Rule & Mean gain (logits) & 95\% CI & Win rate \\
\midrule
\multicolumn{4}{l}{\textit{A. After generation: all ten responses observed}} \\
User-by-prompt router & 0.80 & [0.78, 0.82] & 65\% \\
Prompt router & 0.24 & [0.23, 0.25] & 53\% \\
\midrule
\multicolumn{4}{l}{\textit{B. Before generation, user only: one model per user}} \\
Known tastes & 0.24 & [0.22, 0.25] & 55\% \\
Tastes learned from 10 choices & 0.05 & [0.03, 0.06] & 51\% \\
Tastes learned from 40 choices & 0.14 & [0.12, 0.15] & 53\% \\
\midrule
\multicolumn{4}{l}{\textit{C. Before generation, prompt only: one model per prompt}} \\
Characteristics predicted from the prompt & 0.01 & [0.01, 0.02] & 50\% \\
\midrule
\multicolumn{4}{l}{\textit{D. Before generation, user and prompt}} \\
Known tastes & 0.33 & [0.31, 0.34] & 56\% \\
Tastes learned from 10 choices & 0.09 & [0.08, 0.11] & 51\% \\
Tastes learned from 40 choices & 0.20 & [0.19, 0.22] & 54\% \\
\bottomrule
\end{tabular}

%% file: tables/tab_verbal.tex
\begin{tabular}{lrr}
\toprule
 & Mean gain (logits) & Win rate \\
\midrule
\multicolumn{3}{l}{\textit{A. Steered vs.\ unsteered response, under the tastes of}} \\
The user it was steered for & 1.1 & 68\% \\
The average user & $-0.02$ & 49\% \\
A different user & $-0.01$ & --- \\
\addlinespace
\multicolumn{3}{l}{\textit{B. Gain over the baseline, under the user's tastes}} \\
Prompt router, ten models & 0.15 & 52\% \\
Gemini 2.5 Pro steered for the user & 1.0 & 68\% \\
User-by-prompt router, ten models & 0.77 & 65\% \\
User-by-prompt router, two unsteered responses per model & 0.75 & 65\% \\
Model chosen before generation, then steered & 1.3 & 72\% \\
User-by-prompt router, with steered responses & 1.6 & 77\% \\
\bottomrule
\end{tabular}

%% file: tables/tab_sample.tex
\begin{tabular}{lr@{\qquad}lr}
\toprule
Sample & Battles & Outcome & Share \\
\midrule
Both releases, all languages & 241,768 & A wins & 34\% \\
English & 128,784 & B wins & 34\% \\
Complete features & 128,318 & Tie & 17\% \\
Models seen at least 50 times (estimation sample) & 128,317 & Both bad & 15\% \\
Training split & 115,396 &  &  \\
Test split & 12,921 &  &  \\
\bottomrule
\end{tabular}

%% file: tables/tab_fit.tex
\begin{tabular}{lrrr}
\toprule
Specification & Parameters & $\Delta$ log-lik., training & $\Delta$ log-lik., test \\
\midrule
\multicolumn{4}{l}{\textit{A. Standard logits}} \\
Model fixed effects only & 106 & --- & $-860$ \\
Style and judged score & 129 & $-2{,}200$ & $-210$ \\
\midrule
\multicolumn{4}{l}{\textit{B. Mixed logits, accuracy bins with normal tastes}} \\
Bins & 193 & $-960$ & $-110$ \\
Bins and the value of both bad ($\kappa$) & 196 & $-760$ & $-96$ \\
\midrule
\multicolumn{4}{l}{\textit{C. Mixed logits, judged-score loss $L(s)$ and value of both bad}} \\
No accuracy heterogeneity ($\beta^A_i=1$) & 199 & $-320$ & $-36$ \\
Two types ($\beta^A_i=0$ or a constant) & 200 & $-78$ & $-5.1$ \\
Normal $\beta^A_i$, negative weights allowed & 200 & $-2.8$ & 2.0 \\
Censored normal $\beta^A_i$ (main specification) & 200 & 0 & 0 \\
Three latent classes & 203 & 180 & 18 \\
\bottomrule
\end{tabular}

%% file: tables/tab_recovery.tex
\begin{tabular}{lrrrr}
\toprule
 & True & Mean & Median & At most 0.02 \\
\midrule
Length & 0.30 & 0.30 & 0.30 & 0\% \\
Ordered lists & 0.11 & 0.12 & 0.12 & 8\% \\
Unordered lists & 0.15 & 0.15 & 0.15 & 4\% \\
Headers & 0.01 & 0.08 & 0.06 & 19\% \\
Bold spans & 0.26 & 0.26 & 0.26 & 0\% \\
Emoji & 0.99 & 0.97 & 0.98 & 0\% \\
Table & 0.78 & 0.77 & 0.79 & 0\% \\
Flesch reading ease & 0.19 & 0.18 & 0.18 & 0\% \\
Positive tone & 0.05 & 0.19 & 0.17 & 9\% \\
Negative tone & 0 & 0.32 & 0.24 & 9\% \\
Assertive hedging & 0.41 & 0.42 & 0.43 & 0\% \\
Tentative hedging & 0.30 & 0.32 & 0.33 & 2\% \\
Link & 1.5 & 1.5 & 1.5 & 0\% \\
Intelligence index & 0.02 & 0.05 & 0.04 & 30\% \\
Context window & 0.56 & 0.55 & 0.55 & 0\% \\
Reasoning model & 0.63 & 0.61 & 0.62 & 0\% \\
\midrule
No weight on accuracy, $\Phi(-b/\sigma^A)$ & 0.16 & 0.16 & 0.17 & --- \\
\bottomrule
\end{tabular}

%% file: tables/tab_interactions.tex
\begin{tabular}{lrrr}
\toprule
Feature & $\times$ PLS 1 & $\times$ PLS 2 & $\times$ PLS 3 \\
\midrule
Accuracy weight ($\theta$) & $-0.54$ (0.02) & $-0.12$ (0.02) & 0.14 (0.02) \\
Both bad ($\kappa$) & 0.15 (0.03) & $-0.04$ (0.03) & $-0.16$ (0.03) \\
Length & 0.16 (0.01) & 0.26 (0.01) & 0.06 (0.01) \\
Emoji & 0 (0.02) & 0.02 (0.02) & $-0.02$ (0.02) \\
Table & 0.14 (0.03) & $-0.04$ (0.04) & 0.21 (0.03) \\
Flesch reading ease & $-0.01$ (0.02) & 0.07 (0.02) & 0.03 (0.02) \\
Positive tone & 0.03 (0.01) & 0 (0.01) & 0.02 (0.01) \\
Assertive hedging & 0.01 (0.01) & $-0.06$ (0.01) & $-0.01$ (0.01) \\
Tentative hedging & $-0.03$ (0.02) & 0 (0.02) & $-0.05$ (0.02) \\
\bottomrule
\end{tabular}

%% file: tables/tab_intel.tex
\begin{tabular}{lrrrrr}
\toprule
 & Index & Index bins & Quadratic & ECI & Parameters \\
\midrule
Intelligence index / 10 & 0.12 (0.02) & --- & 0.10 (0.09) & --- & --- \\
Index bin: mid & --- & 0.14 (0.06) & --- & --- & --- \\
Index bin: high & --- & 0.29 (0.07) & --- & --- & --- \\
(Index / 10)$^2$ & --- & --- & 0 (0.01) & --- & --- \\
ECI / 10 & --- & --- & --- & 0.24 (0.04) & --- \\
ln(parameters) & --- & --- & --- & --- & 0.08 (0.02) \\
Context window & $-0.11$ (0.02) & $-0.10$ (0.02) & $-0.11$ (0.02) & $-0.20$ (0.02) & $-0.11$ (0.02) \\
Reasoning model & $-0.05$ (0.07) & 0.06 (0.07) & $-0.05$ (0.08) & $-0.11$ (0.08) & 0.14 (0.05) \\
\midrule
Accuracy loss, score 4 & 0.76 (0.02) & 0.77 (0.02) & 0.77 (0.02) & 0.76 (0.03) & 0.77 (0.02) \\
Accuracy loss, score 7 & 0.29 (0.02) & 0.29 (0.02) & 0.29 (0.02) & 0.29 (0.03) & 0.29 (0.02) \\
No weight on accuracy & 0.16 (0.01) & 0.17 (0.01) & 0.17 (0.01) & 0.18 (0.01) & 0.16 (0.01) \\
\midrule
Battles & 128,317 & 128,317 & 128,317 & 82,415 & 128,317 \\
$\Delta$ test NLL per 100 battles & 0 & 0.01 & $-0.02$ & $-0.02$ & $-0.01$ \\
\bottomrule
\end{tabular}

%% file: tables/tab_plainlogit.tex
\begin{tabular}{lrrr}
\toprule
 & Logit, FE only & Logit, style and accuracy & Mixed logit \\
\midrule
Constant & $-0.21$ (0.13) & 1.0 (0.08) & 1.2 (0.09) \\
Intelligence index / 10 & 0.25 (0.03) & 0.11 (0.02) & 0.12 (0.02) \\
Context window (M tokens) & 0.04 (0.03) & 0.02 (0.02) & $-0.11$ (0.02) \\
Reasoning model & 0.10 (0.11) & $-0.02$ (0.06) & $-0.05$ (0.07) \\
$R^2$ & 0.58 & 0.40 & 0.39 \\
Model fixed effects & 105 & 105 & 105 \\
$\Delta$ test log-likelihood vs. mixed logit & $-860$ & $-210$ & 0 \\
\bottomrule
\end{tabular}

%% file: tables/tab_exante_full.tex
\begin{tabular}{llrrrr}
\toprule
Router & Information used & Gain & 95\% CI & \% of R1 & Win rate \\
\midrule
R1 user-by-prompt router & realized responses, true taste & 0.80 & [0.78, 0.82] & 100 & 65\% \\
R2 prompt router & realized responses & 0.24 & [0.23, 0.25] & 30 & 53\% \\
\midrule
A0 true taste, one model per user & true taste & 0.24 & [0.22, 0.25] & 30 & 55\% \\
A1 learned taste, $k=10$ & 10 past choices & 0.05 & [0.03, 0.06] & 6.0 & 51\% \\
A1 learned taste, $k=40$ & 40 past choices & 0.14 & [0.12, 0.15] & 17 & 53\% \\
A2 classifier, $k=40$ & 40 past choices & 0.10 & [0.09, 0.12] & 13 & 52\% \\
\midrule
B1 ridge to utility & prompt embedding & 0 & [$-0.01$, 0] & $-0.3$ & 50\% \\
B2 network to utility & prompt embedding & 0 & [0, 0] & 0 & 50\% \\
B3 ridge to characteristics & prompt embedding & 0.01 & [0.01, 0.02] & 1.5 & 50\% \\
B4 network to characteristics & prompt embedding & 0 & [$-0.01$, 0.01] & 0 & 50\% \\
\midrule
C1 true taste + B3 & true taste, prompt embedding & 0.33 & [0.31, 0.34] & 41 & 56\% \\
C2 learned taste + B3, $k=10$ & 10 past choices, prompt embedding & 0.09 & [0.08, 0.11] & 12 & 51\% \\
C2 learned taste + B3, $k=40$ & 40 past choices, prompt embedding & 0.20 & [0.19, 0.22] & 25 & 54\% \\
\bottomrule
\end{tabular}

%% file: tables/tab_exante_pred.tex
\begin{tabular}{llrp{1.35in}}
\toprule
Variant & Estimator & Parameters & Fit \\
\midrule
A1 ($k=5$/10/20/40) & approximate posterior mean of $\nu_i$ & 17 per user & taste recovery 0.18 / 0.23 / 0.31 / 0.39 \\
A2 ($k=5$/10/20/40) & multinomial logistic to model id & 220 & label accuracy 0.48 / 0.49 / 0.52 / 0.57 \\
B1 & ridge, embedding to utility (10 targets) & 30,720 & $R^2$ 0.24, routing 0.05 \\
B2 & network, embedding to utility & 1,707,274 & $R^2$ 0.25, routing 0.04 \\
B3 & ridge, embedding to 14 characteristics $\times$ 10 models & 430,080 & $R^2$ 0.32 (characteristics), 0.27 (utility), routing 0.06 \\
B4 & network, embedding to characteristics & 1,740,684 & $R^2$ 0.33 (characteristics), 0.27 (utility), routing 0.05 \\
\bottomrule
\end{tabular}

%% file: tables/tab_bestofn.tex
\begin{tabular}{lrrrrrr}
\toprule
& \multicolumn{2}{c}{User-by-prompt router} & \multicolumn{2}{c}{Prompt router} & \multicolumn{2}{c}{Models vs.\ Gemini} \\
\cmidrule(lr){2-3}\cmidrule(lr){4-5}\cmidrule(lr){6-7}
Answers & Gemini & Models & Gemini & Models & User-by-prompt & Prompt \\
\midrule
2 & 55\% (0.24) & 55\% (0.28) & 53\% (0.16) & 51\% (0.11) & 50\% & 49\% \\
3 & 57\% (0.34) & 58\% (0.42) & 54\% (0.22) & 52\% (0.15) & 51\% & 48\% \\
5 & 58\% (0.44) & 61\% (0.59) & 55\% (0.27) & 53\% (0.19) & 53\% & 48\% \\
10 & 61\% (0.55) & 65\% (0.81) & 56\% (0.32) & 54\% (0.25) & 55\% & 48\% \\
\midrule
All 19 & --- & 68\% (0.96) & --- & 56\% (0.38) & --- & --- \\
\bottomrule
\end{tabular}

%% file: tables/tab_verbal_models.tex
\begin{tabular}{lrrrrr}
\toprule
& \multicolumn{3}{c}{Steered vs. own unsteered} & \multicolumn{2}{c}{vs. always Gemini} \\
\cmidrule(lr){2-4}\cmidrule(lr){5-6}
Policy & Mean gain & Win rate & Placebo gain & Mean gain & Win rate \\
\midrule
Steering GPT-4.1 & 1.2 & 71\% & 0.01 & 0.76 & 63\% \\
Steering Llama 4 Maverick & 1.2 & 69\% & 0.08 & 0.22 & 54\% \\
Steering Qwen3 235B & 1.1 & 69\% & $-0.10$ & 0.32 & 56\% \\
Steering Gemini 2.5 Pro & 1.0 & 68\% & $-0.11$ & 1.0 & 68\% \\
Steering GPT-4o & 0.97 & 67\% & 0.01 & 0.21 & 55\% \\
Steering DeepSeek V3 & 0.95 & 67\% & $-0.15$ & 0.59 & 60\% \\
\midrule
Prompt router, six models &  &  &  & 0.11 & 52\% \\
User-by-prompt router, six models &  &  &  & 0.60 & 62\% \\
Prompt router, ten models &  &  &  & 0.15 & 52\% \\
User-by-prompt router, ten models &  &  &  & 0.77 & 65\% \\
Steering, best model per user and prompt &  &  &  & 1.5 & 75\% \\
User-by-prompt router, panel and steered &  &  &  & 1.6 & 77\% \\
\bottomrule
\end{tabular}